%% file: template.tex
\documentclass{article}

\usepackage{arxiv}

\usepackage{amsmath} 
\usepackage{oldgerm} 
\usepackage{tabularray}
\usepackage{tabularx} 
\usepackage[hyphens]{url}
\usepackage[colorlinks = true,
            linkcolor = blue,
            urlcolor  = blue,
            citecolor = blue,
            anchorcolor = blue]{hyperref}
\usepackage{booktabs}       
\usepackage{amsfonts}       
\usepackage{nicefrac}       
\usepackage{microtype}      
\usepackage{lipsum}		
\usepackage{graphicx}
\usepackage{doi}

\usepackage[utf8]{inputenc} 
\usepackage[T1]{fontenc} 
\usepackage{subcaption}
\usepackage{caption}

\usepackage{float}
\usepackage{color}
\usepackage{tabularray}
\usepackage{array}
\usepackage{multirow}
\usepackage{ragged2e}
\usepackage{booktabs}
\usepackage{vcell}
\usepackage{subcaption}
\usepackage{amsmath,amssymb,amsfonts}
\usepackage{algorithmic}
\usepackage{graphicx}
\usepackage{textcomp}
\usepackage{booktabs}
\usepackage{multirow}
\usepackage{subcaption}
\usepackage{caption}
\usepackage{siunitx}
\usepackage{pifont}
\usepackage{tipa}
\usepackage{tfrupee}
\usepackage{soul}
\usepackage{adjustbox}
\usepackage{tabularx,ragged2e,booktabs}

\usepackage{graphicx}

\title{Camouflage is all you need: Evaluating and Enhancing Language Model Robustness Against Camouflage Adversarial Attacks}

\author{ \href{https://orcid.org/0000-0003-2165-0144}{\includegraphics[scale=0.06]{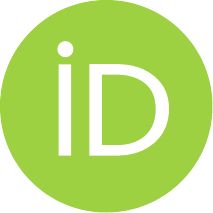}\hspace{1mm}{\'A}lvaro Huertas-Garc{\'i}a}
\\
	Department of Computer Systems Engineering\\
	Universidad Polit{\'e}cnica de Madrid\\
	Madrid, Spain \\
	\texttt{alvaro.huertas.garcia@upm.es} \\
	\And
	\href{https://orcid.org/0000-0002-0800-7632}{\includegraphics[scale=0.06]{orcid.pdf}\hspace{1mm}Alejandro Mart{\'i}n} \\
	Department of Computer Systems Engineering\\
	Universidad Polit{\'e}cnica de Madrid\\
	Madrid, Spain \\
	\texttt{alejandro.martin@upm.es} \\
	\And
	\href{https://orcid.org/0000-0003-4127-5505}{\includegraphics[scale=0.06]{orcid.pdf}\hspace{1mm}Javier Huertas-Tato} \\
	Department of Computer Systems Engineering\\
	Universidad Polit{\'e}cnica de Madrid\\
	Madrid, Spain \\
	\texttt{javier.huertas.tato@upm.es} \\	
	\And
	\href{https://orcid.org/0000-0002-5051-3475}{\includegraphics[scale=0.06]{orcid.pdf}\hspace{1mm}David Camacho} \\
	Department of Computer Systems Engineering\\
	Universidad Polit{\'e}cnica de Madrid\\
	Madrid, Spain \\
	\texttt{david.camacho@upm.es} \\	
}

\renewcommand{\shorttitle}{Camouflage is all you need}

\hypersetup{
pdftitle={Camouflage is all you need: Evaluating and Enhancing Language Model Robustness Against Camouflage Adversarial Attacks},
pdfsubject={q-bio.NC, q-bio.QM},
pdfauthor={David S.~Hippocampus, Elias D.~Striatum},
pdfkeywords={First keyword, Second keyword, More},
}

\begin{document}
\maketitle

\begin{abstract}
Adversarial attacks represent a substantial challenge in Natural Language Processing (NLP). This study undertakes a systematic exploration of this challenge in two distinct phases: vulnerability evaluation and resilience enhancement of Transformer-based models under adversarial attacks.

In the evaluation phase, we assess the susceptibility of three Transformer configurations—encoder-decoder, encoder-only, and decoder-only setups—to adversarial attacks of escalating complexity across datasets containing offensive language and misinformation. Encoder-only models manifest a performance drop of 14\% and 21\% in offensive language detection and misinformation detection tasks respectively. Decoder-only models register a 16\% decrease in both tasks, while encoder-decoder models exhibit a maximum performance drop of 14\% and 26\% in the respective tasks.

The resilience-enhancement phase employs adversarial training, integrating pre-camouflaged and dynamically altered data. This approach effectively reduces the performance drop in encoder-only models to an average of 5\% in offensive language detection and 2\% in misinformation detection tasks. Decoder-only models, occasionally exceeding original performance, limit the performance drop to 7\% and 2\% in the respective tasks. Encoder-decoder models, albeit not surpassing the original performance, can reduce the drop to an average of 6\% and 2\% respectively.

These results suggest a trade-off between performance and robustness, albeit not always a strict one, with some models maintaining similar performance while gaining robustness. Our study, which includes the adversarial training techniques used, has been incorporated into an open-source tool that will facilitate future work in generating camouflaged datasets. Although our methodology shows promise, its effectiveness is subject to the specific camouflage technique and nature of data encountered, emphasizing the necessity for continued exploration.
\end{abstract}


\section{Introduction}

The rapid advancement of Artificial Intelligence (AI) and its increasing ubiquity in various domains have underscored the importance of ensuring the robustness and reliability of machine learning models~\cite{steps_toward_robustness_2017, adversarial_train_imporve_2021}. A particular area of concern lies in the field of Natural Language Processing (NLP)~\cite{adversarial_review_2019}, where Transformer-based language models have proven to be very effective in tasks ranging from sentiment analysis and text classification to question answering~\cite{transformer_review_2023, transformers_sota_2020}.

One of the novel and key challenges in NLP relates to adversarial attacks, where subtle modifications are made to input data to fool the models into making incorrect predictions~\cite{adversarial_review_2019, black_box_attack_2017}. A relevant example of this concept is the use of word camouflage techniques. For instance, the phrase "Word camouflage" can be subtly altered to ``W0rd cam0uflage" or ``VV0rd cam0ufl4g3." While these changes are often unnoticeable to a human reader, they can lead a machine learning model to misinterpret or misclassify the input~\cite{adversarial_nlp_2017}. This raises substantial ethical issues around misinformation dissemination, content evasion, and the potential for AI systems to be exploited for malicious intent~\cite{worldhealth2021infodemic,  martin2021factercheck}.

Real-world implications of these attacks are increasingly evident, with numerous instances of adversarial attacks compromising online content moderation systems, leading to the spread of harmful content~\cite{paolo_survey}. Existing methods to counter such adversarial attacks have limitations and often focus on post-attack detection~\cite{pdf_malware_evasion_2016, adversarial_post_attack_2020}, failing to proactively prevent the occurrence of such attacks. Additionally, these methods often struggle with text data which is discrete in nature, making it challenging to apply perturbation methods that were originally designed for continuous data like images~\cite{adversarial_train_imporve_2021, adversarial_review_2019,evaluate_cnn_image_robustness_2020}. 

This study introduces a comprehensive methodology to evaluate and strength the resilience of Transformer-based language models to camouflage adversarial attacks. We examine the vulnerability and robustness of distinct Transformer configurations—encoder-decoder, encoder-only, and decoder-only—in two use-cases involving offensive language and false information datasets.

The research adopts a proactive defense strategy of adversarial training, which incorporates camouflaged data into the training phase. This is accomplished either by statically camouflaging the dataset or by dynamically altering it during training. A key contribution is the development of an open-source tool\footnote{Omitted for anonymity reasons.} that generates various versions of camouflaged datasets, offering a range of difficulty levels, camouflage techniques, and proportions of camouflaged data.

The evaluation of model vulnerability is based on an unbiased methodology, drawing from significant literature references~\cite{bridge_gap_leet_2005, romero_wordcamo_2021, Blashki2005GameGG, fuchs__2013, craenen_leet_cheatsheet} and using AugLy~\cite{papakipos2022augly}, a data augmentation library, for external validation. This approach helps ensure that our assessment of model weakness accurately reflects real-world content evasion and misclassification techniques.

In addressing the the current gaps in the field~\cite{adversarial_review_2019}, our research provides insights into three critical challenges—perceivability (the extent to which adversarial changes are noticeable), transferability (the ability of an attack to be effective across different models), and automation (the ability to generate adversarial examples automatically) of camouflaged adversarial examples.

Preliminary results indicate considerable susceptibility across various Transformer configurations, with performance drops reaching up to 14\% in offensive language detection and 26\% in misinformation detection tasks, highlighting the salient requirement for robustness augmentation.

In addition to identifying these vulnerabilities, the research explores the enhancement of models against such threats. Employing adversarial training methodologies that amalgamate both pre-camouflaged and dynamically altered data, the study uncovers promising avenues for resilience improvement. However, the effectiveness of the approach is influenced by several variables, including the complexity of the camouflage technique employed and the nature and distribution of data encountered by the model, thus emphasising the urgent need for further research in this domain.

The research paper is organised as follows: Section 2 reviews pertinent literature, illuminating the gaps in existing methodologies that this research endeavours to address. Section 3 provides a detailed outline of the methodology employed to develop word camouflage adversarial attacks, while Section 4 explicates the procedure undertaken for enhancing and evaluating the resilience of Transformer models. Section 5 presents empirical findings from each stage of the study, evidencing the impact of adversarial attacks on naive Transformer models, alongside the efficacy of the adversarial fine-tuning approach implemented. Finally, Section 6 offers an in-depth discussion of the research's implications, ethical considerations, and limitations, concluding with recommendations for future exploration.

\section{Background and Related Work}

This section will cover the various aspects of adversarial attacks within the field of Natural Language Processing (NLP). Topics discussed will include the definition of adversarial attacks, the taxonomy of attacks, the measurement of perturbations, and the evaluation metrics for attack effectiveness. It will also explore the impact of these attacks on deep learning models in NLP tasks and potential solutions and defenses against these attacks.

\subsection{Conceptual Aspects of Adversarial Attacks in NLP}

Adversarial attacks constitute a significant challenge to deep learning models, as they introduce minimal, often imperceptible, changes to input data with the aim of triggering incorrect model outputs~\cite{adversarial_review_2019, evaluate_cnn_image_robustness_2020, goodfellow2015explaining}.

Adversarial attacks fall into two main categories: white-box and black-box attacks. In white-box attacks, attackers possess complete access to the model's architecture, parameters, loss functions, activation functions, and input/output data, allowing them to craft more precise and effective attacks~\cite{adversarial_review_2019, cheng2020seq2sick, michel-etal-2019-evaluation, alzantot-etal-2018-generating}. Conversely, black-box attacks operate under limited knowledge, with attackers only having access to the model's inputs and outputs. Despite the constraints, these attacks can still induce incorrect model outputs and are more reflective of real-world scenarios~\cite{adversarial_review_2019, black_box_attack_2017}.

Perturbation measurements, controlling the magnitude and direction of alterations, and evaluation metrics, like accuracy and F1 score, allow for a standardized comparison of different adversarial attacks' impact on model performance~\cite{adversarial_review_2019, evaluate_cnn_image_robustness_2020, adversarial_nlp_2017, adversarial_post_attack_2020}.

Although adversarial threats persist, several defenses have been proposed, including adversarial training, defensive distillation, feature squeezing, and input sanitization~\cite{goodfellow2015explaining, adversarial_review_image_2018, adversarial_review_ml_2018}. However, these defenses remain susceptible to more sophisticated attacks~\cite{adversarial_review_2019}.

These challenges are particularly relevant for Transformer models given their broad usage in NLP tasks, making the investigation of their susceptibility to adversarial attacks and potential defenses highly significant.

\subsection{Transformers and Their Significance}

Transformer models have revolutionized Natural Language Processing (NLP), offering exceptional performance across a variety of tasks and thereby becoming a leading model in the field~\cite{transformers_sota_2020, transformer_review_nlp__2021, transformer_review_2023}. Unlike their predecessors, Recurrent Neural Networks (RNNs) and Convolutional Neural Networks (CNNs), Transformers employ attention mechanisms, allowing for simultaneous input processing and resulting in enhanced performance on context-dependent tasks~\cite{rnn_architecture_1986, CNN_NLP, dnn_nlp_review_2020, vaswani2017attention}.

Transformer models can adopt multiple configurations. Encoder-decoder models are suitable for tasks like machine translation, requiring the generation of output sequences based on input understanding~\cite{transformer_review_2023, t5-model}. Encoder-only models, such as BERT, focus on input representations, ideal for tasks like text classification~\cite{devlin-etal-2019-bert}. Decoder-only models, like GPT, facilitate sequence generation based on a given context, useful in text generation or language translation tasks~\cite{gpt_2_2019, decoder-model}.

In the face of adversarial attacks, the complexity of Transformer models presents both challenges and opportunities~\cite{adversarial_review_2019}. While their superior performance can lead to vulnerabilities, their configurational flexibility could provide potential defense strategies, making the understanding of their architecture essential in the context of adversarial attacks.

\subsection{Previous Works}
The exploration of Deep Neural Networks (DNNs) for text data has lagged behind image data, though recent research has begun to reveal DNN vulnerabilities in text-based tasks~\cite{adversarial_review_2019,adversarial_review_image_2018}. Studies have shown that carefully crafted adversarial text samples can manipulate DNN-based classifiers~\cite{adversarial_review_2019}.

Significant work, such as those by Liang et al.\cite{Liang_2018} and Cheng et al.\cite{cheng2020seq2sick}, has been done in the white-box attack domain, demonstrating the feasibility of misleading DNN text classifiers with adversarial samples. In another study, Blohm et al.\cite{blohm-etal-2018-comparing} focused on comparing the robustness of convolutional and recurrent neural networks via a white-box approach. Guo et al.\cite{guo-etal-2021-gradient} offered a white-box attack method for transformer models, but their approach neglected token insertions and deletions, compromising the naturalness of the adversarial examples.

Black-box adversarial attacks have gained attention recently. Noteworthy work by Maheshwary et al.\cite{maheshwary2020context} presented an attack strategy that generated high-quality adversarial examples with a high success rate. The work of Gil et al.\cite{gil2019whitetoblack} distilled white-box attack experience into a neural network to expedite adversarial example generation. Similarly, Li et al.~\cite{Li2022TextualAA} proposed a black-box adversarial attack for named entity recognition tasks, demonstrating the importance of understanding black-box attacks for developing robust models.

Adversarial attacks have profound real-world implications, ranging from security risks in autonomous driving~\cite{deng2020analysis} to societal structures like voter dynamics~\cite{adversarial_attack_real_word_online_2022} and social networks~\cite{FChen2022AnAM, Yin2023A2S2GNNRG}. These examples underscore the importance of addressing adversarial attack vulnerabilities across domains, emphasizing naturalness, black-box evaluation, and flexibility.

\subsection{Adressing Open Issues}
Several challenges persist in the field of adversarial attacks on text data~\cite{adversarial_review_2019}. Primarily, creating textual adversarial examples is complex due to the need to maintain syntax, grammar, and semantics, which makes fooling natural language processing (NLP) systems difficult~\cite{adversarial_train_imporve_2021, adversarial_review_2019, evaluate_cnn_image_robustness_2020}. Transferability needs a deeper understanding across different architectures and datasets, and automating adversarial example generation remains difficult. Additionally, with the advent of new architectures like generative models and those with attention mechanisms, their vulnerability to adversarial attacks needs to be explored.

To address these issues, the paper proposes an approach focusing on new architectures, transferability, and automation in black-box adversarial attacks. The researchers explore a range of Transformer model configurations, study transferability across different datasets and architectures, and design an automated method for generating adversarial examples. They rely on literature references for unbiased evaluation~\cite{bridge_gap_leet_2005, romero_wordcamo_2021, Blashki2005GameGG, fuchs__2013, craenen_leet_cheatsheet} and AugLy~\cite{papakipos2022augly} for external validation. The focus on black-box adversarial attacks helps simulate real-world scenarios, contributing to enhanced security in areas like social networks and content moderation.

\section{Methodology}
This section starts with Phase I, emphasizing the `Camouflaging Techniques' and `Camouflage Difficulty Levels' in the development and evaluation of adversarial attacks (subsections \ref{sec:tecniques} and \ref{sec:levels}). Phase II then explores the `Fine-tuning Approaches' (subsection \ref{sec:fine-tune}) to enhance model resilience against word camouflage attacks, followed by an `External Validation' to ensure robustness (subsection \ref{sec:external-val}).

\begin{figure}[tpb]
\centering
\includegraphics[width=\linewidth]{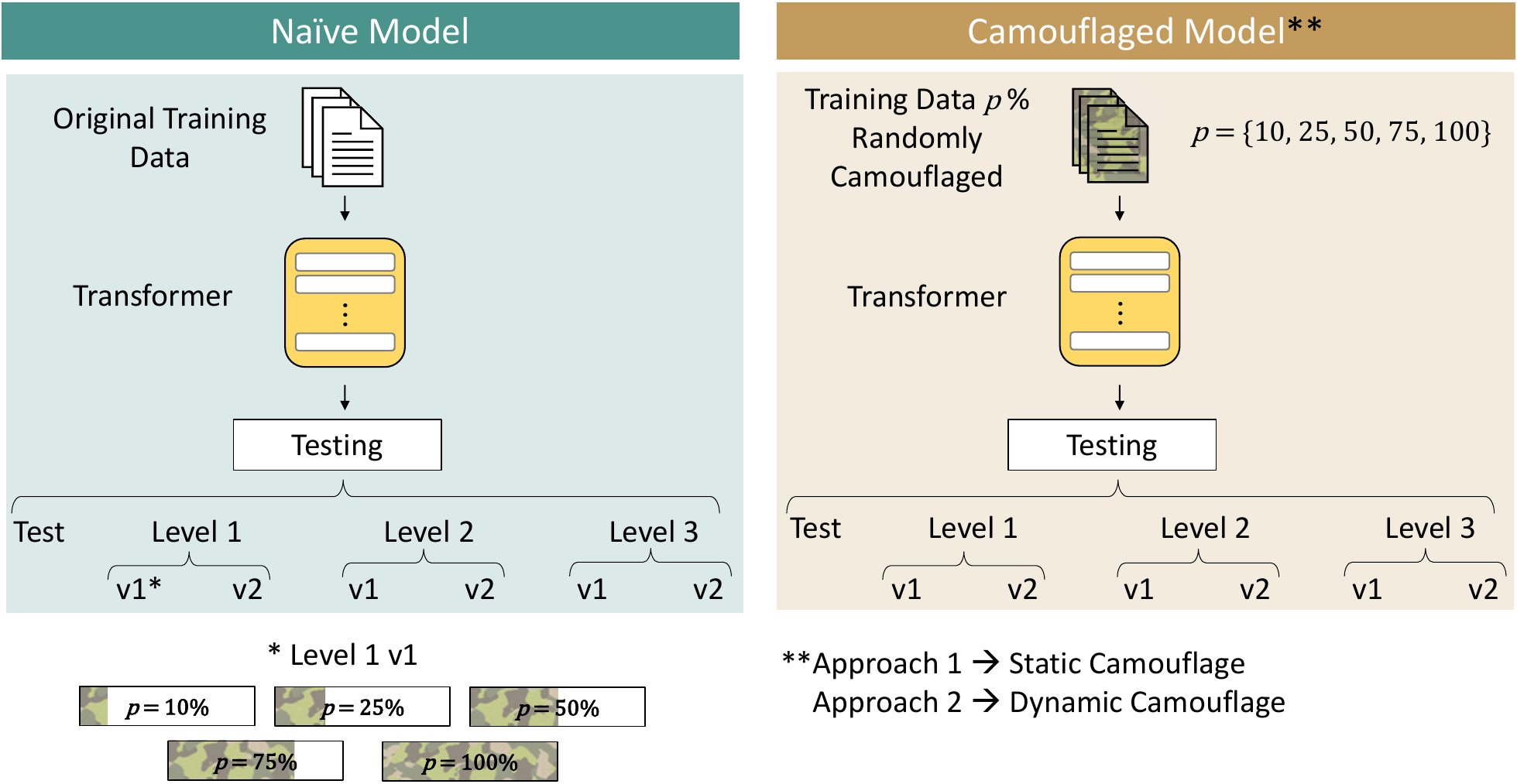}
\caption{Methodology for training and evaluating Transformed models to assess word camouflage robustness. Left side: naive model trained on original dataset and tested on various versions (Te, C-Te-Lvl1/2/3) with different camouflaged keywords and percentages (p) of modified instances as highlighted in (*). Right side: Camouflaged models trained on data with mixed random level modifications, developed for different percentages (p) of modifications. (**) Two approaches highlighted: Approach 1 with pre-camouflaged training data and Approach 2 with on-the-fly data camouflage during training.}
\label{fig:diagram_methodology}
\end{figure}

\begin{table}[t!]
\centering
\begin{minipage}[b]{0.5\textwidth}

\input{table_parameters}

\end{minipage}
\hspace{1mm}
\centering
\begin{minipage}[b]{0.4\textwidth}
  \input{table_training_parameters}
\end{minipage}
\end{table}

\begin{figure}[htpb]
\centering
\includegraphics[width=\linewidth]{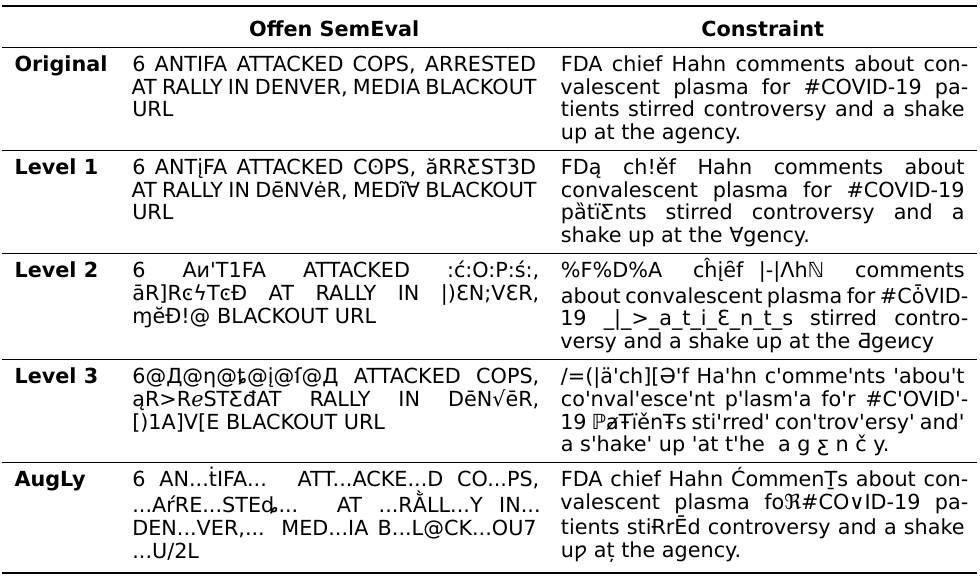}
\caption{Comparison of original and camouflaged text examples from the Offen SemEval 2019 and Constraint datasets. The table presents examples of three levels of camouflage by the tool introduced in this study, as well as an example from the AugLy library for external validation with unseen modifications. Each level represents increasing complexity of camouflage.}
\label{table:data-example}
\end{figure}

\subsection{Phase I: Evaluating the Impact of Word Camouflage on Transformer Models}

\subsubsection{Camouflaging Techniques}
\label{sec:tecniques}

The study employs the ``pyleetspeak" Python package, developed based on recent research~\cite{huertasgarcía2022countering}, for generating realistic adversarial text samples. This package applies three distinct text camouflage techniques, prioritizing semantic keyword extraction over random selection, thereby providing a more realistic representation of word camouflage threats. Selected for their relevance and varied complexity in adversarial attacks, these techniques offer a robust evaluation of model resilience. These techniques are:

\begin{itemize}
    \item Leetspeak: This technique involves substituting alphabet characters with visually analogous symbols or numbers, creating changes that can range from basic to highly intricate. As an illustration, Offensive" could be altered to 0ff3ns1v3" with vowel and specific consonant replacements.

    \item Punctuation Insertion: This method alters text by introducing punctuation symbols to create visually similar character strings. Punctuation can be inserted at hyphenation points or between any two characters. For example, ``fake news" could be camouflaged as ``f-a-k-e n-e-w-s".
    
    \item Syllable Inversion: This technique, less commonly used, camouflages words by rearranging their syllables. For instance, ``Methodology" could be altered to ``Me-do-tho-lo-gy" by inverting the syllables in the word.
    
\end{itemize}

The use of these techniques helps to perform a thorough investigation of the Transformer models' resilience to different kinds of adversarial attacks. 
\subsubsection{Camouflage Difficulty Levels}
\label{sec:levels}
Adversarial camouflage attacks in this study are systematically varied across three parameters: complexity level, word camouflage ratio (amount of words camouflaged within a data instance), and instance camouflage ratio (the proportion of camouflaged instances within a test dataset). This approach enables a comprehensive assessment of the models' susceptibility to these attacks. Each complexity level involves specific parameters, as detailed in Table~\ref{table_parameters}. Examples of each level are depicted in Table~\ref{table:data-example}.

\begin{itemize}
    \item \textbf{Level 1} serves as a starting point and is quite readable and understandable. It introduces minor changes to the text, primarily focusing on simple character substitutions, such as replacing every vowel with a number or similar symbol.

    \item \textbf{Level 2} increases the complexity of modifications. It introduces extended and complex character substitutions, punctuation injections, and simple word inversions. It extends substitutions to both vowels and consonants, introducing also readable symbols from other alphabets that closely resemble regular alphabet characters.

    \item \textbf{Level 3} is the most complex tier. It combines techniques from the previous two levels but intensifies the use of punctuation marks, introduces more character substitutions, and incorporates inversions, and mathematical symbols are also incorporated making the text even more challenging to comprehend. 

\end{itemize}

For each complexity level, two versions are produced. These versions, known as `v1' and `v2', represent different word camouflage ratios, camouflaging 15\% and 65\% of words per text, respectively. These ratios were derived from the statistical distribution of text lengths in the employed datasets.

These three complexity levels emulate real-world adversarial camouflage attacks, facilitating a methodical progression from low to high complexity, and thus, allowing a detailed evaluation of model robustness against incrementally challenging adversarial scenarios.

Regarding, instance camouflage ratio, as depicted in Figure~\ref{fig:diagram_methodology}, the tests systematically introduce varying levels and percentages of camouflage into the data instances. Starting from an original test set (Te), 30 additional tests are generated across three difficulty levels, each with two word camouflage ratio versions (v1 and v2), and five different percentages of camouflaged instances (10\%, 25\%, 50\%, 75\%, and 100\%). This structure, totalling 31 tests, provides a robust framework for assessing the impact of increasing prevalence and complexity of camouflage techniques. Further discussion on this methodology is covered in Section \ref{sec:eval-robust}.

This systematic approach allows a granular evaluation of word camouflage adversarial attacks, providing insights into their implications and potential countermeasures.


\subsection{Phase II: Improving Resilience Against Word Camouflage Attacks}
\subsubsection{Fine-tuning Approaches}
\label{sec:fine-tune}

In our study, as depicted in Figure~\ref{fig:diagram_methodology}, we train and evaluate two distinct sets of models for each task: `Naive Models' and `Camouflaged Models'.

The Naive Models represent a baseline approach. These models are trained on the original, unaltered datasets, thus reflecting traditional methods of natural language processing model training. Once training is complete, these models are then evaluated across the 31 different tests we defined earlier, which incorporate varying degrees and proportions of word camouflage.

On the other hand, the Camouflaged Models integrate adversarial data during their training phase. These models are trained on datasets that are modified to include instances of word camouflage, drawn randomly from the three levels of difficulty we previously described. Notably, a distinct Camouflaged Model is trained for each possible percentage of camouflaged data instances (10\%, 25\%, 50\%, 75\%, and 100\%). This approach allows us to investigate how the distribution or frequency of camouflaged instances within the training data may affect the model's robustness and performance.

Including all three levels of camouflage in the training of these models aims to enhance their ability to generalize across a variety of adversarial conditions. However, given the potential impact of camouflage frequency on training, we maintain the percentage of camouflaged instances consistent for each individual model.

Furthermore, for each Camouflaged Model, we employ two different fine-tuning strategies, as depicted in Figure~\ref{fig:diagram_methodology}:

\begin{itemize}
    \item \textbf{Static Modification}: This approach involves camouflaging the training dataset prior to model training. The benefit of this method lies in its simplicity and predictability, as the dataset remains consistent throughout the training process. However, this may limit the model's ability to adapt to new or varying types of camouflage not represented in the initial dataset.

    \item \textbf{Dynamic Modification}: In contrast, this method involves camouflaging the training dataset `on-the-fly', introducing changes during the training process itself. This allows the model to be exposed to a wider variety and unpredictability of camouflage techniques, potentially improving its ability to generalize and adapt. The trade-off, however, is a more complex and computationally demanding training process.
\end{itemize}

Together, these different training and evaluation strategies provide a comprehensive approach to understanding the susceptibility of natural language processing models to word camouflage, and the potential strategies for improving their robustness. The parameters employed to the fine-tuning approaches can be found at ~\ref{table:training-params}.

\subsubsection{External Validation}
\label{sec:external-val}

The AugLy library~\cite{papakipos2022augly}, developed by Meta AI, is used as an instrument of external validation in order to ensure the objectiveness and applicability of our evaluations. By generating an auxiliary test dataset, we can re-affirm our evaluations on model resilience against word camouflage. This library extends a distinctive approach of random text modifications, including letter substitutions with analogous Unicode or non-Unicode characters, punctuation insertions and font alterations. While these manipulations are based on random selection, instead of keyword extraction as in our methodology, they nonetheless simulate potential evasion techniques, providing a valuable counterpoint to our stratified camouflage techniques.

AugLy diverge from the `pyleetspeak' package implemented in our research, which select word to be camouflaged based on semantic importance. Furthermore, AugLy exhibits a limited degree of flexibility in terms of user control and omits features such as word inversion and modification tracking.

Nevertheless, as an instrument of external validation, AugLy assumes a crucial role, especially when compared with the three-tier complexity levels. Comparative analysis with AugLy facilitates the identification of potential model vulnerabilities and ensures unbiased results, ensuring a comprehensive defense strategy. This approach guarantees rigorous model evaluation and improves real-world performance resilience.

\section{Experimental Setup}

This section outlines the critical components of the research design. Subsection \ref{sec:models}  details the specific architectures employed in the study, while the 'Data' subsection \ref{sec:data} describes the datasets used for adversarial sample creation and model training and evaluation. Lastly, subsection \ref{sec:eval-robust} establishes the evaluation metrics and procedures for determining model resilience against adversarial attacks.

\subsection{Transformer Models}
\label{sec:models}
In this investigation, three distinct Transformer models are employed, each representing a unique configuration: encoder-only, decoder-only, and encoder-decoder. This selection facilitates the comparison of performance and resilience to adversarial attacks under diverse operational conditions.

BERT (bert-base-uncased)~\cite{devlin-etal-2019-bert} serves as the encoder-only model. Renowned for its wide application and being among the most downloaded models on HuggingFace\footnote{\href{https://huggingface.co/models?pipeline\_tag=fill-mask\&sort=downloads}{https://huggingface.co/models?pipeline\_tag=fill-mask\&sort=downloads}}, BERT provides an ideal case for examining the potential susceptibility of prevalent models to adversarial onslaughts and word camouflage. Pretrained with the primary objectives of Masked Language Modeling (MLM) and Next Sentence Prediction (NSP), BERT boasts approximately 110M parameters.

The mBART model (mbart-large-50)~\cite{mbart_model_2020} functions as the encoder-decoder paradigm. An extension of the original mBART model, it supports 50 languages for multilingual machine translation models. Its "Multilingual Denoising Pretraining" objective introduces noise to the input text, potentially augmenting its robustness against adversarial attacks. Developed by Facebook, this model encapsulates over 610M parameters.

Lastly, Pythia (pythia-410m-deduped)~\cite{biderman2023pythia}, forming part of EleutherAI's Pythia Scaling Suite, is harnessed. With its training on the Pile~\cite{gao2020pile}, a dataset recognized for its diverse range of English texts, it offers a fitting choice for analyzing model resilience to the often biased and offensive language pervasive on the internet. Comprising 410M parameters, it is well-equipped for this study.

\subsection{Data}
\label{sec:data}

The study utilizes two primary datasets, OffensEval~\cite{zampieri-etal-2019-semeval} and Constraint~\cite{constraint_2021}, representing distinct aspects of online behavior. OffensEval, part of the SemEval suite, consists of over 14,000 English tweets focusing on offensive language in social media. On the other hand, Constraint pertains to the detection of fake news related to COVID-19 across various social platforms, comprising a collection of 10,700 manually annotated posts and articles.

These datasets underscore the significance of combatting offensive language and misinformation, especially during critical times like a pandemic. They provide a solid foundation for examining the resilience of Transformer models to adversarial attacks and gauging the efficacy of camouflage techniques in evading detection.

To ensure data quality and reliability, several preprocessing steps were undertaken. These include eliminating duplicates, filtering out instances with fewer than three characters, preserving the original text case, and verifying the balance of the binary classes across both datasets. This balance verification is pivotal to ensure a fair assessment of camouflage techniques across different classes and avoid bias introduced by class imbalances.

Post-preprocessing, the datasets were divided into training, validation, and test sets, ensuring a comprehensive and balanced evaluation. For OffensEval and Constraint, the training set comprised 11,886 and 6,420 instances, respectively, with appropriate allocations for validation and testing.

\subsection{Evaluating Robustness}
\label{sec:eval-robust}

For the evaluation of model performance, the F1-macro score is used. Despite the datasets being balanced, the choice of F1-macro score offers a more rigorous measure that enables the effective isolation of the impact of word camouflage techniques on performance and minimizes any potential bias due to class distribution.

A comprehensive experimental setup is implemented, encompassing 31 internal tests and an external test from AugLy. This detailed framework facilitates an in-depth assessment of how increasing complexity of camouflage techniques influence model performance and the practical application of the results.

Model robustness is assessed by the degree of performance reduction when models are evaluated on camouflaged test datasets at varied levels, relative to their performance on the original test dataset. The degree of performance reduction serves as a systematic measure of model resilience against adversarial attacks and illuminates how model performance deteriorates as difficulty levels increase and percentage of modified data escalates.

\section{Experiment Results and Discussion}
\subsection{Phase I: Assessing Transformer Models’ Susceptibility to Word Camouflage}
\label{sec:phase1}

\begin{table}[t]
    \caption{Encoder-Only Transformer models original performance and performance reduction across camouflage levels, and AugLy. Models exceeding Naive ones are marked (*), with the least impacted by camouflage highlighted in bold.}
    \centering
    \begin{subtable}{0.49\linewidth}
        \centering
        \input{table_bert_offen_weakness}
        \caption{Encoder-only OffensEval comparative performance}
        \label{table:bert_weakness_offen}
    \end{subtable}
    \hspace{1mm}
    \begin{subtable}{0.49\linewidth}
        \centering
        \input{table_bert_constraint_weakness}
        \caption{Encoder-only Constraint comparative performance}
        \label{table:bert_weakness_constraint}
    \end{subtable}

\label{table:bert_weakness}
\end{table}

\begin{table}[ht]
    \caption{
Decoder-Only Transformer models original performance and performance reduction across camouflage levels, and AugLy. Models exceeding Naive ones are marked (*), with the least impacted by camouflage highlighted in bold.}
\vspace{2mm}
    \centering
    \begin{subtable}{0.49\linewidth}
        \centering
        \input{table_pythia_offen_weakness}
        \caption{Decoder-only OffensEval comparative performance}
        \label{table:pythia_offen_waekness}
    \end{subtable}
    \hspace{1mm}
    \begin{subtable}{0.49\linewidth}
        \centering

\input{table_pythia_constraint_weakness}
        \caption{Decoder-only Constraint comparative performance}
        \label{table:pythia_weakness_constraint}
    \end{subtable}
\label{table:pythia_weakness}
\end{table}    
\begin{table}[tpb]
    \caption{
Comparative Performance and Resilience of Encoder-Decoder Transformer Models to Word Camouflage on the Offensive Language Task from (a) OffensEval and (b) Constraint. The 'Test F1-Macro' column indicates the F1-Macro score achieved by each model on the original, non-camouflaged Test dataset. Models outperforming the Naive model are marked with an asterisk (*). The 'Weakness' columns report the percentage reduction in performance on different levels of camouflaged Test datasets compared to the original Test dataset. In these columns, the model with the lowest percentage reduction is highlighted in bold, and the model with the greatest weakness is in italics. The 'AugLy' column indicates model performance on the AugLy camouflaged dataset.}
\vspace{2mm}
    \centering
    \begin{subtable}{0.49\linewidth}
        \centering

\input{table_mbart_offen_weakness}
        \caption{Encoder-Decoder OffensEval comparative performance}
        \label{table:mbart_weakness_offen}
    \end{subtable}
    \hspace{1mm}
    \begin{subtable}{0.49\linewidth}
        \centering

\input{table_mbart_constraint_weakness}
        \caption{Encoder-Decoder Constraint comparative performance}
        \label{table:mbart_weakness_constraint}
    \end{subtable}
\label{table:mbart_weakness}
\end{table}

The initial phase of the study is dedicated to investigating the vulnerability of Naive transformer models to adversarial attacks conducted through word camouflage. These models, having been trained on original datasets without prior exposure to word camouflage, are evaluated using the OffensEval and Constraint tasks. This assessment includes all three transformer configurations - Encoder-only, Decoder-only, and Encoder-Decoder, each being scrutinised independently.

A consistently emerging trend across all tasks and model configurations is the significant reduction in performance of Naive models as the complexity of camouflage levels intensifies.

As an illustration, in the OffensEval task, the Encoder-only Naive model's performance reduces by an average of 8\%, starting from a 2\% decrease at Level 1 and peaking at a 14\% decrease at Level 3 (as depicted in Table \ref{table:bert_weakness_offen}). The Decoder-only Naive models showcase a similar pattern, albeit with a steeper average performance decline of 12\%, reaching a maximum of 16\% at Level 3 (refer to Table \ref{table:pythia_offen_waekness}). Among all configurations, the Encoder-Decoder model exhibits the least drastic performance decline, with an average decrease of 10\% and a maximum decrease of 10\% at Level 3 (refer to Table \ref{table:mbart_weakness_offen}).

The same pattern is observed in the Constraint task, with the performance of the Naive models progressively deteriorates with the escalation of camouflage levels. The Encoder-only model, as presented in Table \ref{table:bert_weakness_constraint}, endures an average performance reduction of 12\% across all levels, culminating in a 21\% reduction at Level 3. In addition to this, the Decoder-only and Encoder-Decoder configurations experience average performance reductions of 10\% and 15\% respectively (Tables \ref{table:pythia_weakness_constraint}, \ref{table:mbart_weakness_constraint}).

These findings underscore the profound impact of increasing complexity on model performance, thereby spotlighting the inherent weaknesses that various evasion techniques can exploit. Intriguingly, when examining the performance reduction results for different configurations (refer to Tables \ref{table:bert_weakness}, \ref{table:pythia_weakness}, and \ref{table:mbart_weakness}), it is evident that all models face heightened difficulties at Levels 2 and 3 compared to Level 1. Specifically, Level 3 and v2 (featuring an increased ratio of camouflaged words in each data instance) across all levels prove to be the most challenging.

\begin{figure}[htpb]
\textbf{Encoder-only - OffensEval Results}
\vspace{0.3cm}
\centering

\begin{subfigure}[h]{0.48\linewidth}
\includegraphics[width=\linewidth]{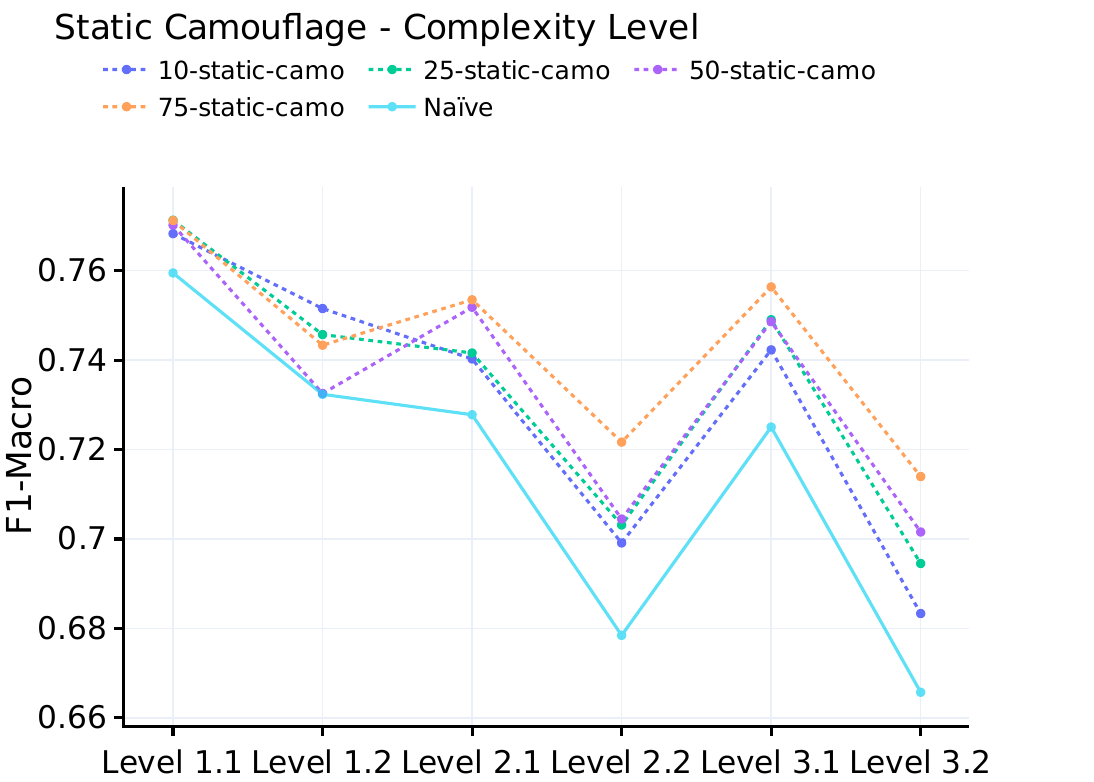}
\caption{}
\label{fig:bert_static_offen_level}
\end{subfigure}
\begin{subfigure}[h]{0.48\linewidth}
\includegraphics[width=\linewidth]{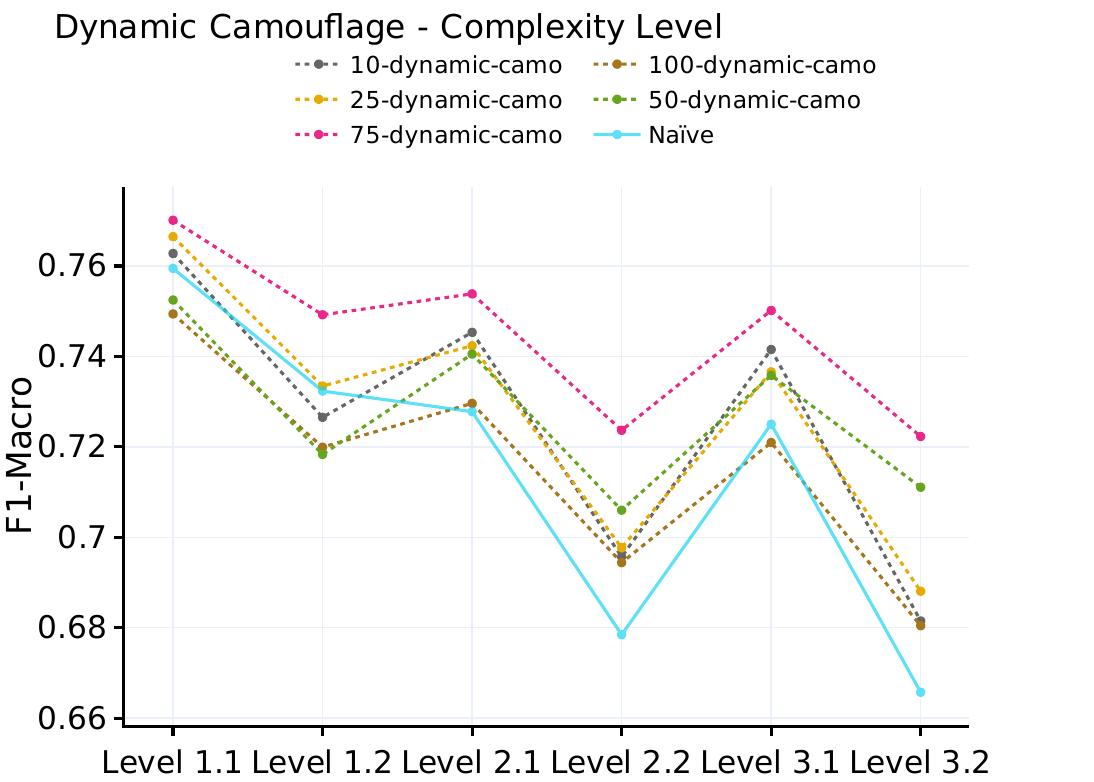}
\caption{}
\label{fig:bert_dynamic_offen_level}
\end{subfigure}

\vspace{0.3cm}

\begin{subfigure}[h]{0.48\linewidth}
\includegraphics[width=\linewidth]{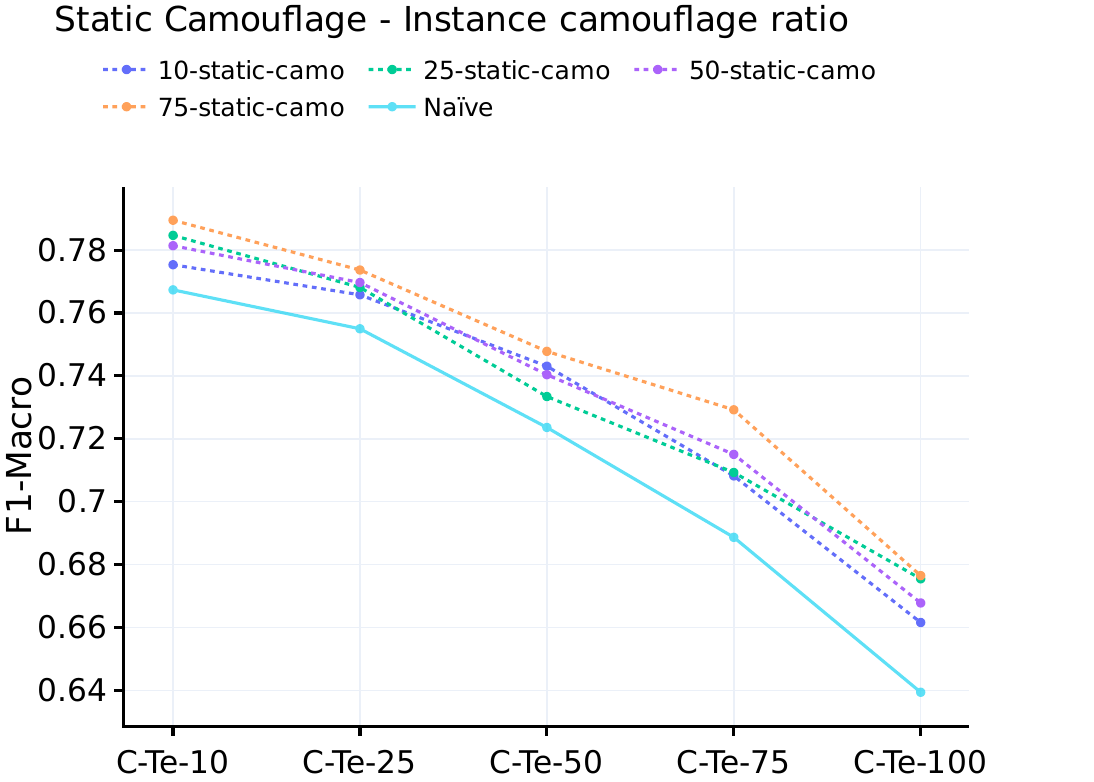}
\caption{}
\label{fig:bert_offen_static_percentage}
\end{subfigure}
\begin{subfigure}[h]{0.48\linewidth}
\includegraphics[width=\linewidth]{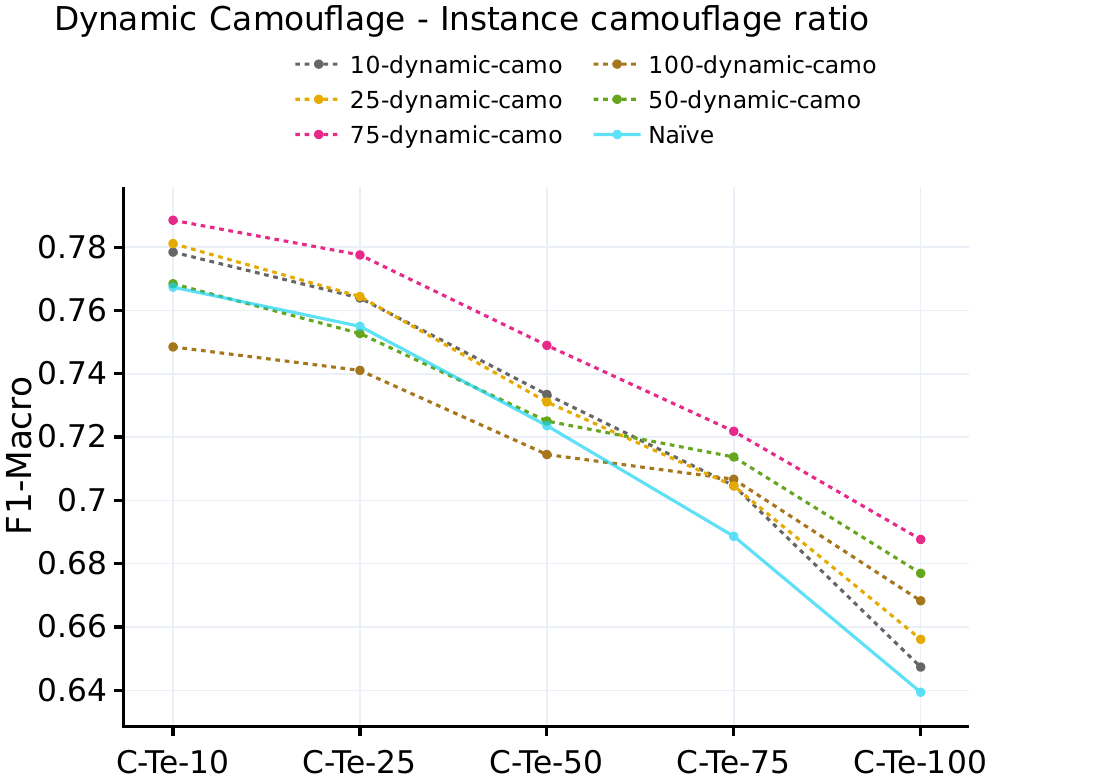}
\caption{}
\label{fig:bert_offen_dynamic_percentage}
\end{subfigure}

\caption{Comprehensive performance comparison of fine-tuned Encoder-only models against naive models in the Offensive Language task from OffensEval under various conditions. (a) Performance of Pre-camouflaged Models across different levels. (b) Performance of Var-camouflaged Models across different levels. (c) Performance of Pre-camouflaged Models across different camouflage percentages. (d) Performance of Var-camouflaged Models across different camouflage percentages.}
\label{fig:bert_offen_plot}
\end{figure}

\begin{figure}[htpb]
\textbf{Encoder-only - Constraint Results}
\vspace{0.3cm}
\centering

\begin{subfigure}[h]{0.48\linewidth}
\includegraphics[width=\linewidth]{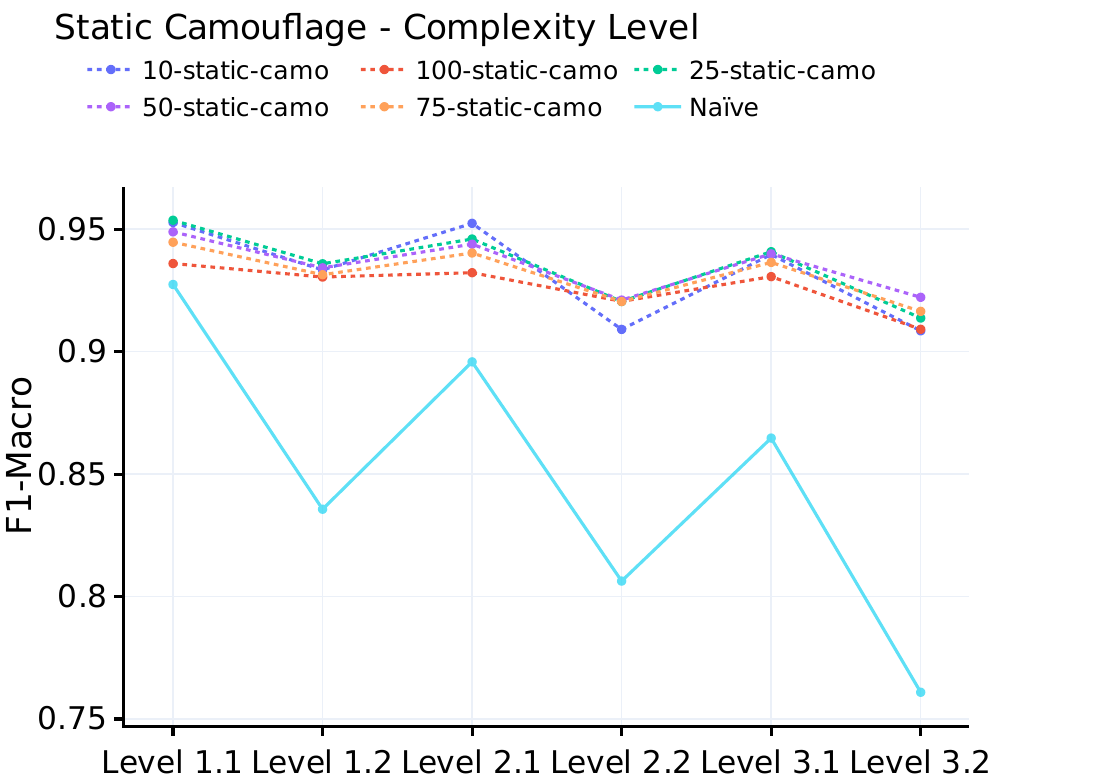}
\caption{}
\label{fig:constraint_per_levels}
\end{subfigure}
\begin{subfigure}[h]{0.48\linewidth}
\includegraphics[width=\linewidth]{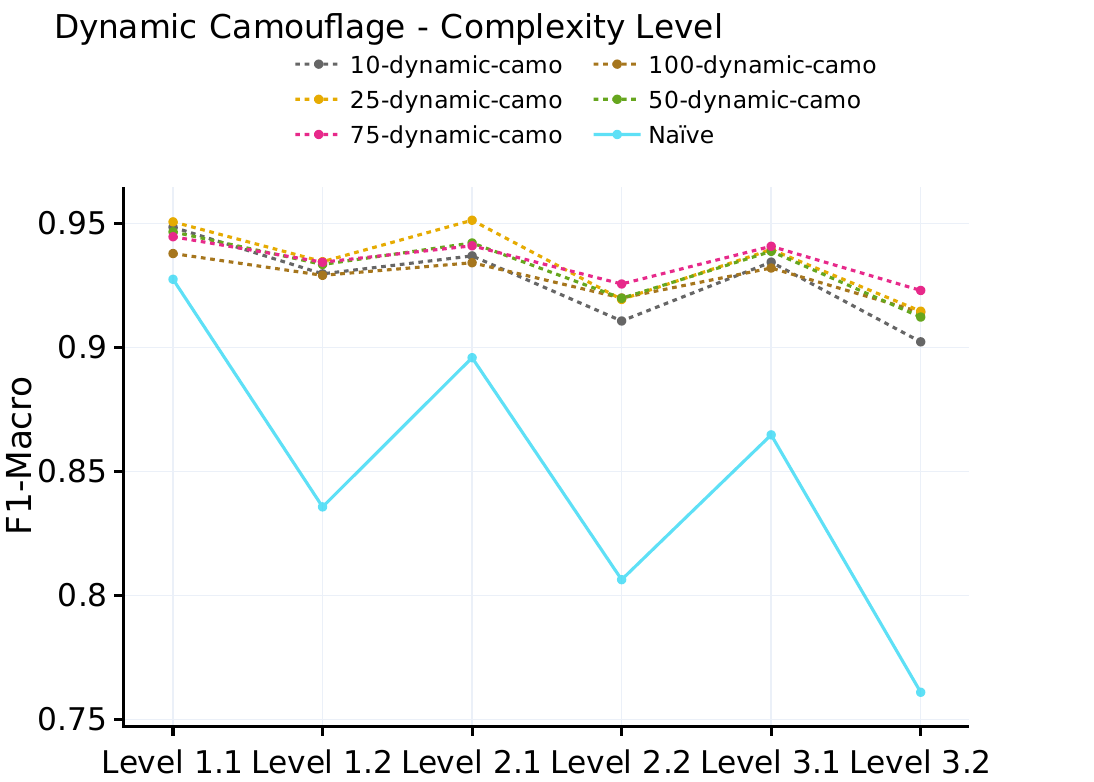}
\caption{}
\label{fig:constraint_var_levels}
\end{subfigure}

\vspace{0.3cm}

\begin{subfigure}[h]{0.48\linewidth}
\includegraphics[width=\linewidth]{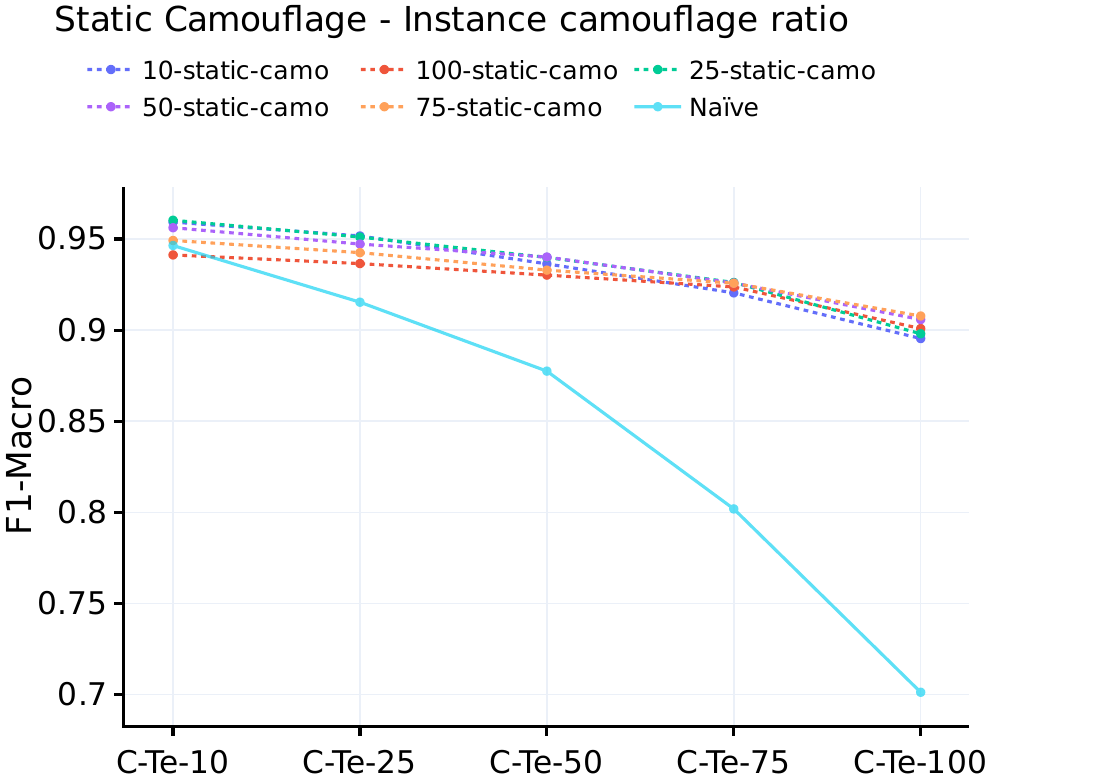}
\caption{}
\label{fig:constraint_pre_percentage}
\end{subfigure}
\begin{subfigure}[h]{0.48\linewidth}
\includegraphics[width=\linewidth]{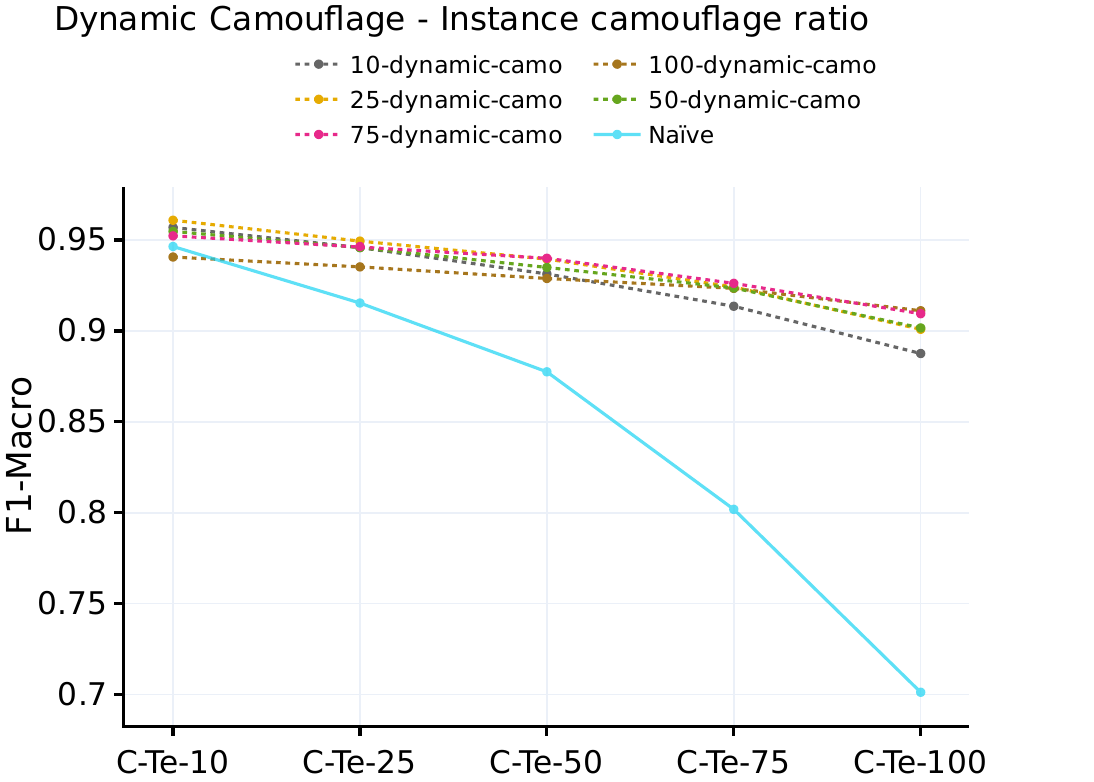}
\caption{}
\label{fig:constraint_var_percentage}
\end{subfigure}

\caption{Comprehensive performance comparison of fine-tuned Encoder-only models against naive models in the False Information Language task from Constraint under various conditions. (a) Performance of Pre-camouflaged Models across different levels. (b) Performance of Var-camouflaged Models across different levels. (c) Performance of Pre-camouflaged Models across different camouflage percentages. (d) Performance of Var-camouflaged Models across different camouflage percentages.}
\label{fig:constraint_plots}
\end{figure}

Particularly, when more camouflaged words are present in a text (represented as v2), the model faces greater challenges than when fewer words are camouflaged (represented as v1). This conclusion is substantiated by the consistently elevated performance reduction percentages witnessed across all v2 levels compared to their v1 counterparts. It suggests that a seemingly straightforward strategy, such as the augmentation of words camouflaged within a text, can substantially compromise model performance.

\begin{figure}[htpb]
\textbf{Decoder-only - OffensEval Results}
\vspace{0.3cm}
\centering

\begin{subfigure}[h]{0.48\linewidth}
\includegraphics[width=\linewidth]{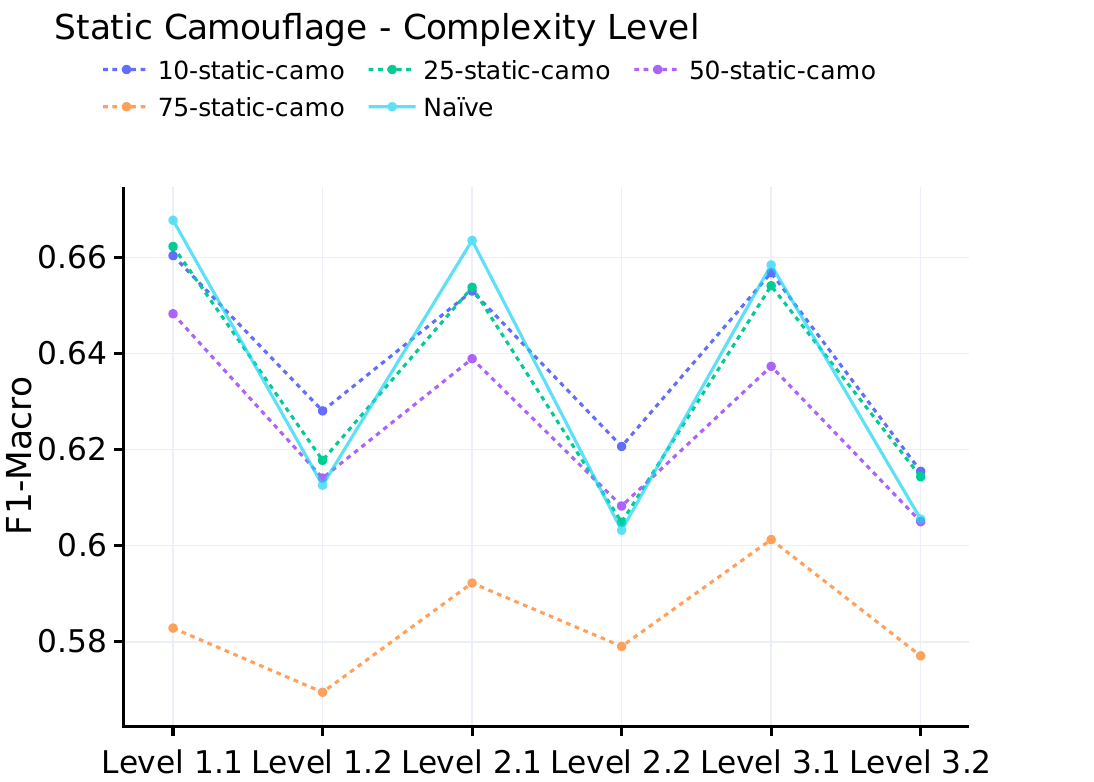}
\caption{}
\label{fig:pythia_offen_per_levels}
\end{subfigure}
\begin{subfigure}[h]{0.48\linewidth}
\includegraphics[width=\linewidth]{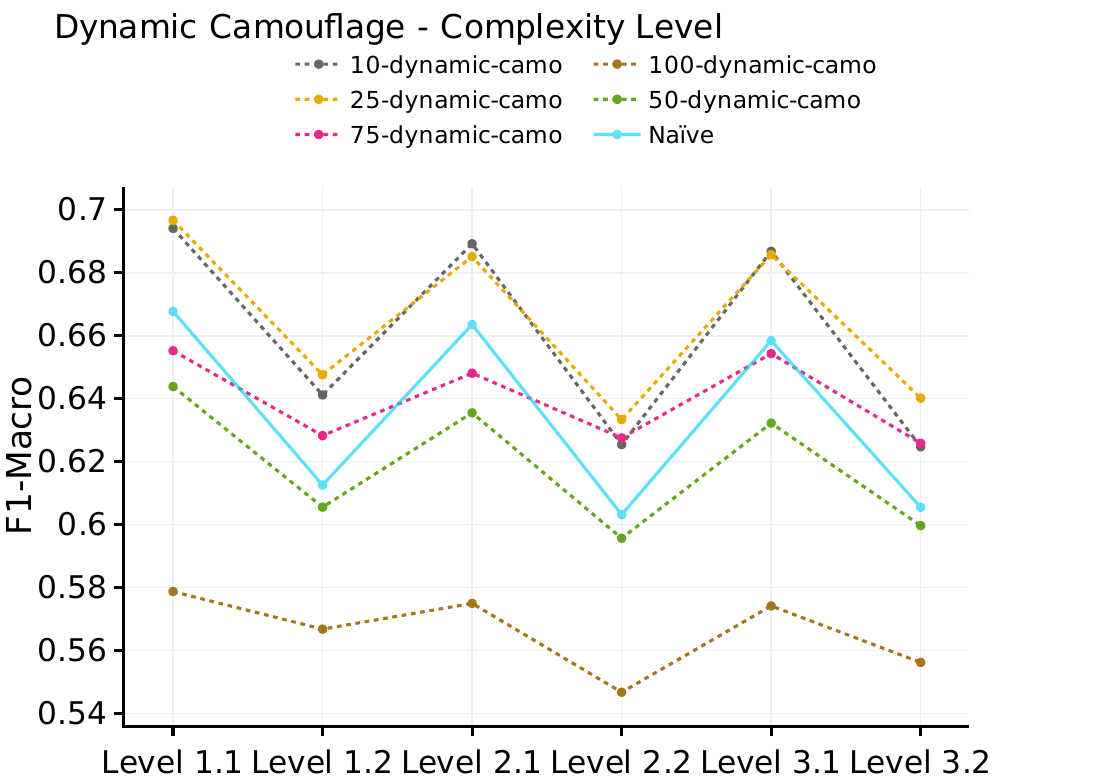}
\caption{}
\label{fig:pythia_offen_var_levels}
\end{subfigure}

\vspace{0.3cm}

\begin{subfigure}[h]{0.48\linewidth}
\includegraphics[width=\linewidth]{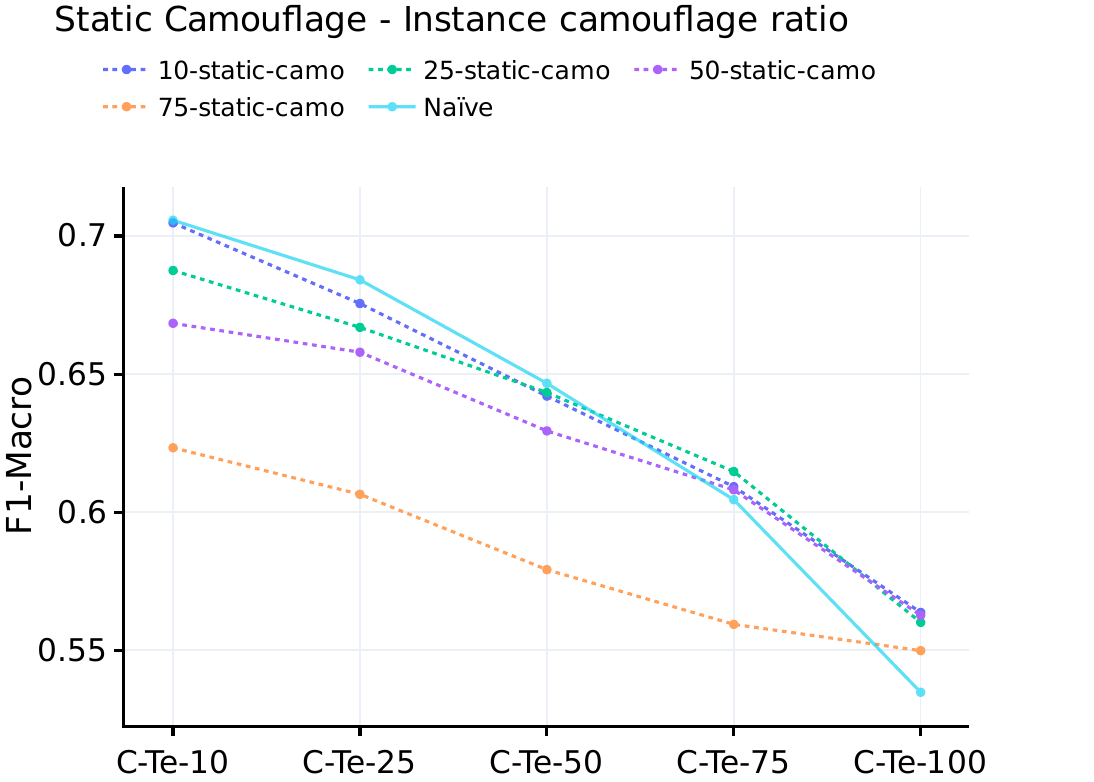}
\caption{}
\label{fig:pythia_offen_pre_percentage}
\end{subfigure}
\begin{subfigure}[h]{0.48\linewidth}
\includegraphics[width=\linewidth]{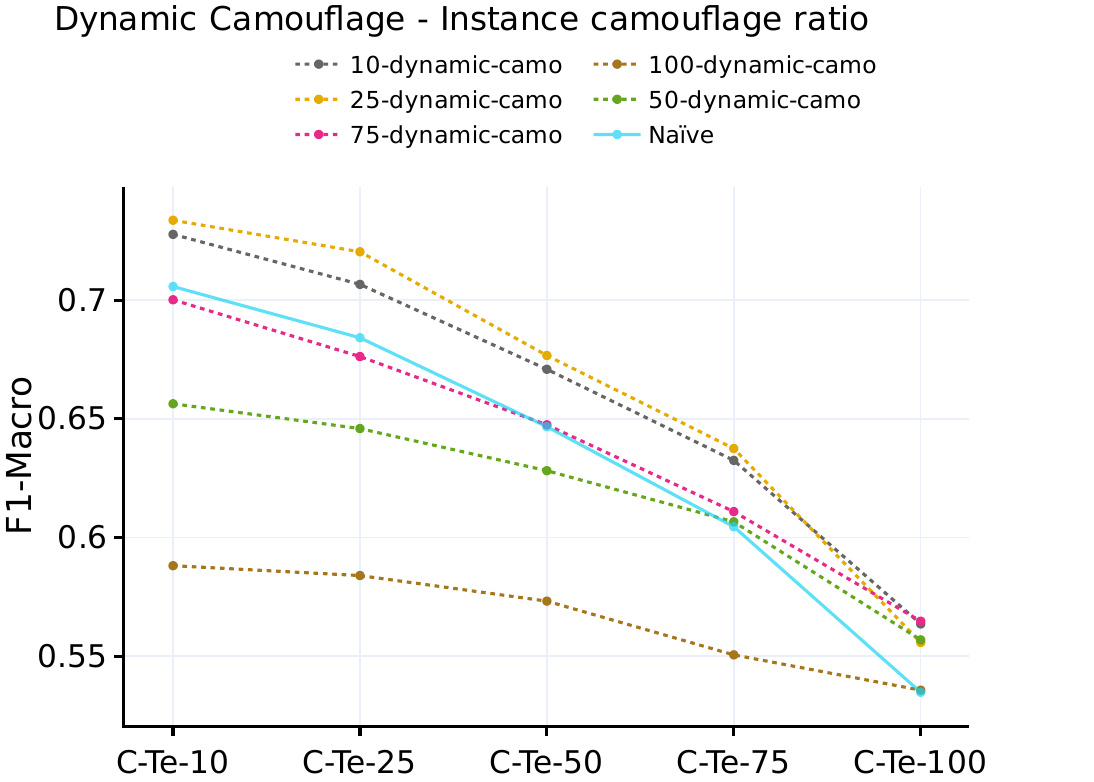}
\caption{}
\label{fig:pythia_offen_var_percentage}
\end{subfigure}

\caption{Comprehensive performance comparison of fine-tuned Decoder-only models against naive models in the Offensive Language task from OffensEval under various conditions. (a) Performance of Pre-camouflaged Models across different levels. (b) Performance of Var-camouflaged Models across different levels. (c) Performance of Pre-camouflaged Models across different camouflage percentages. (d) Performance of Var-camouflaged Models across different camouflage percentages.}
\label{fig:pythia_offen_plots}
\end{figure}

\begin{figure}[tpb]
\textbf{Decoder-only - Constraint Results}
\vspace{0.3cm}
\centering

\begin{subfigure}[h]{0.48\linewidth}
\includegraphics[width=\linewidth]{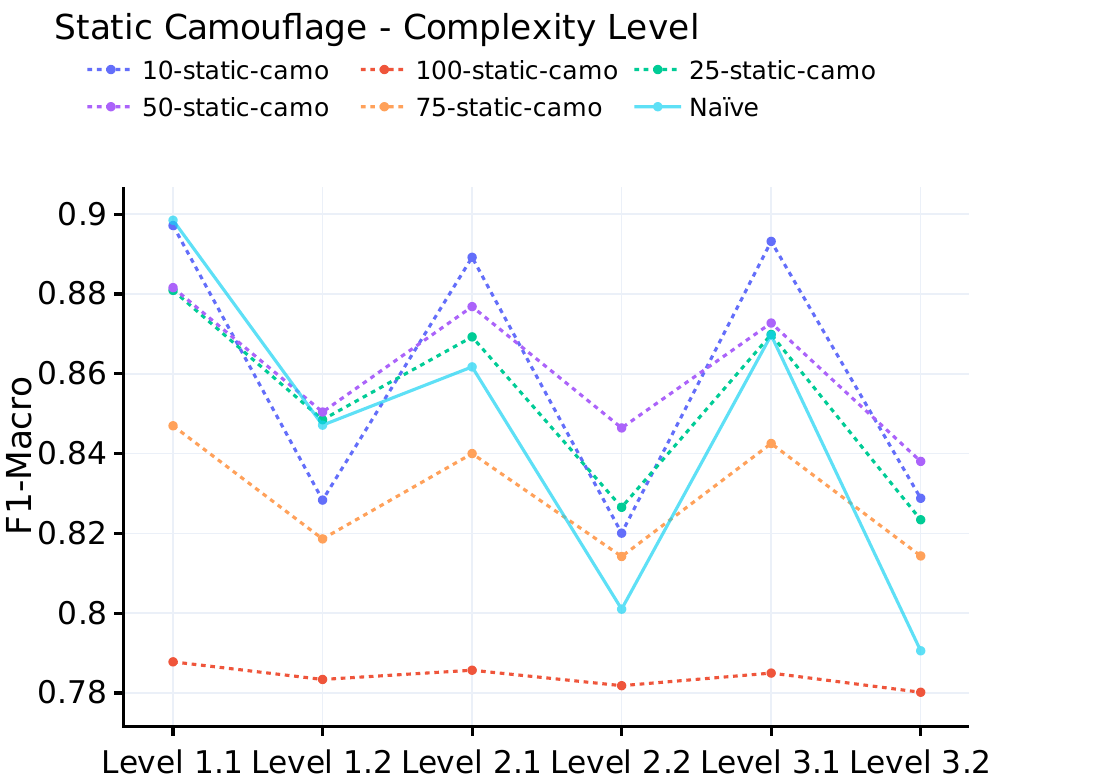}
\caption{}
\label{fig:pythia_constraint_per_levels}
\end{subfigure}
\begin{subfigure}[h]{0.48\linewidth}
\includegraphics[width=\linewidth]{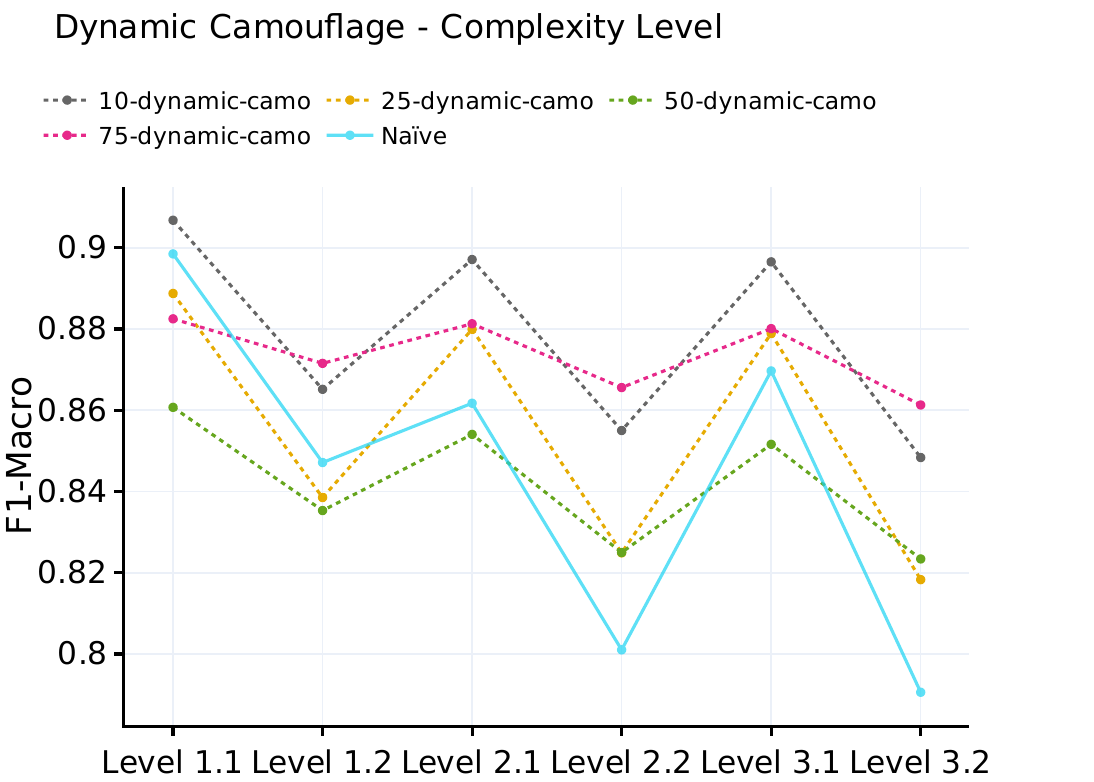}
\caption{}
\label{fig:pythia_constraint_var_levels}
\end{subfigure}

\vspace{0.3cm}

\begin{subfigure}[h]{0.48\linewidth}
\includegraphics[width=\linewidth]{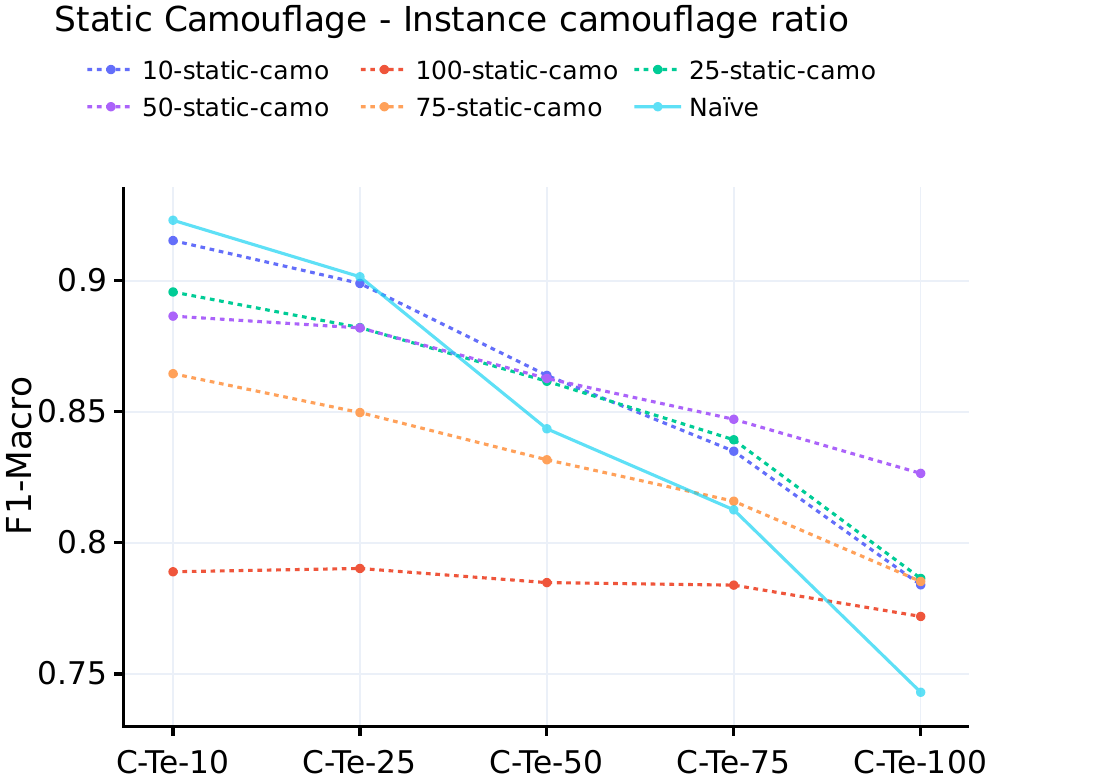}
\caption{}
\label{fig:pythia_constraint_pre_percentage}
\end{subfigure}
\begin{subfigure}[h]{0.48\linewidth}
\includegraphics[width=\linewidth]{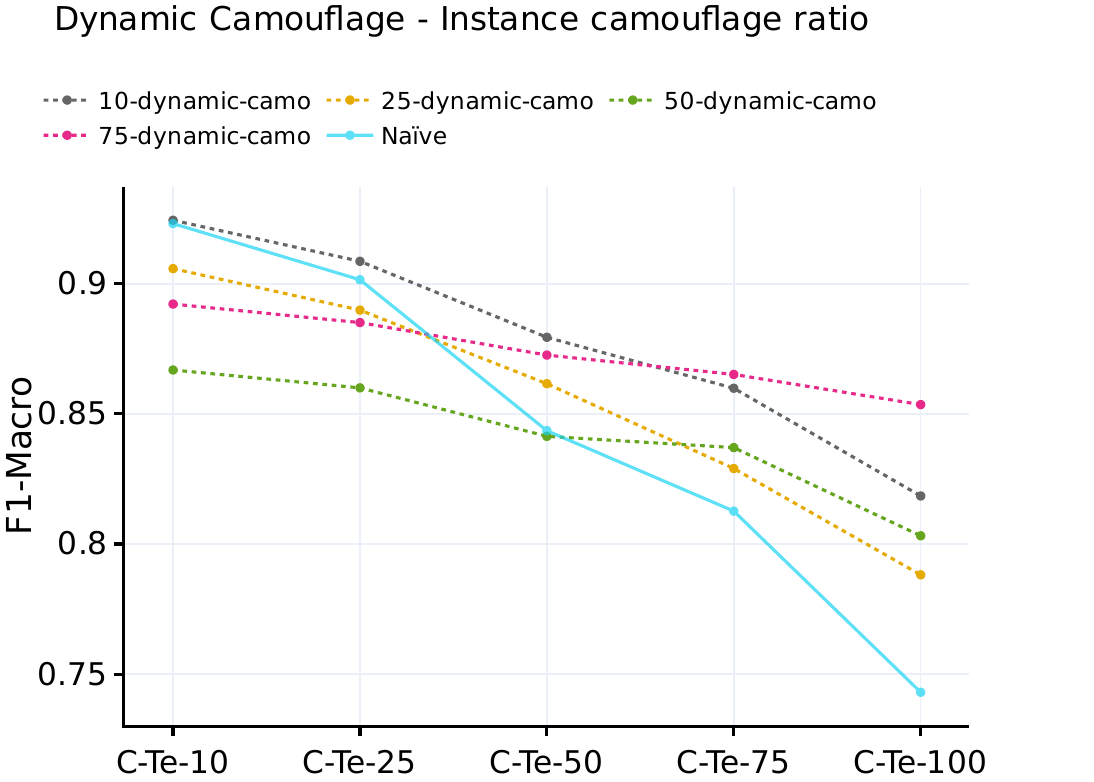}
\caption{}
\label{fig:pythia_constraint_var_percentage}
\end{subfigure}

\caption{Comprehensive performance comparison of fine-tuned Decoder-only models against naive models in the False Information Language task from Constraint under various conditions. (a) Performance of Pre-camouflaged Models across different levels. (b) Performance of Var-camouflaged Models across different levels. (c) Performance of Pre-camouflaged Models across different camouflage percentages. (d) Performance of Var-camouflaged Models across different camouflage percentages.}
\label{fig:pythia_constraint_plots}
\end{figure}

\begin{figure}[tpb]
\textbf{Encoder-Decoder - OffensEval Results}
\vspace{0.3cm}
\centering

\begin{subfigure}[h]{0.48\linewidth}
\includegraphics[width=\linewidth]{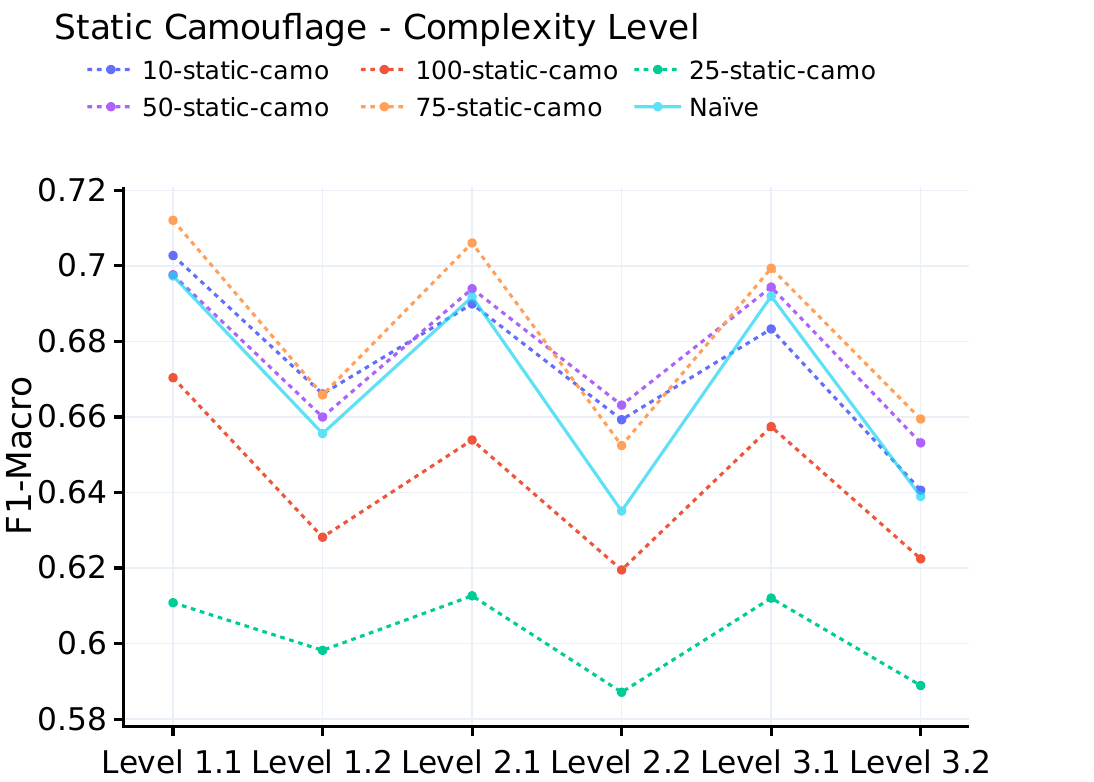}
\caption{}
\label{fig:mbart_offen_per_levels}
\end{subfigure}
\begin{subfigure}[h]{0.48\linewidth}
\includegraphics[width=\linewidth]{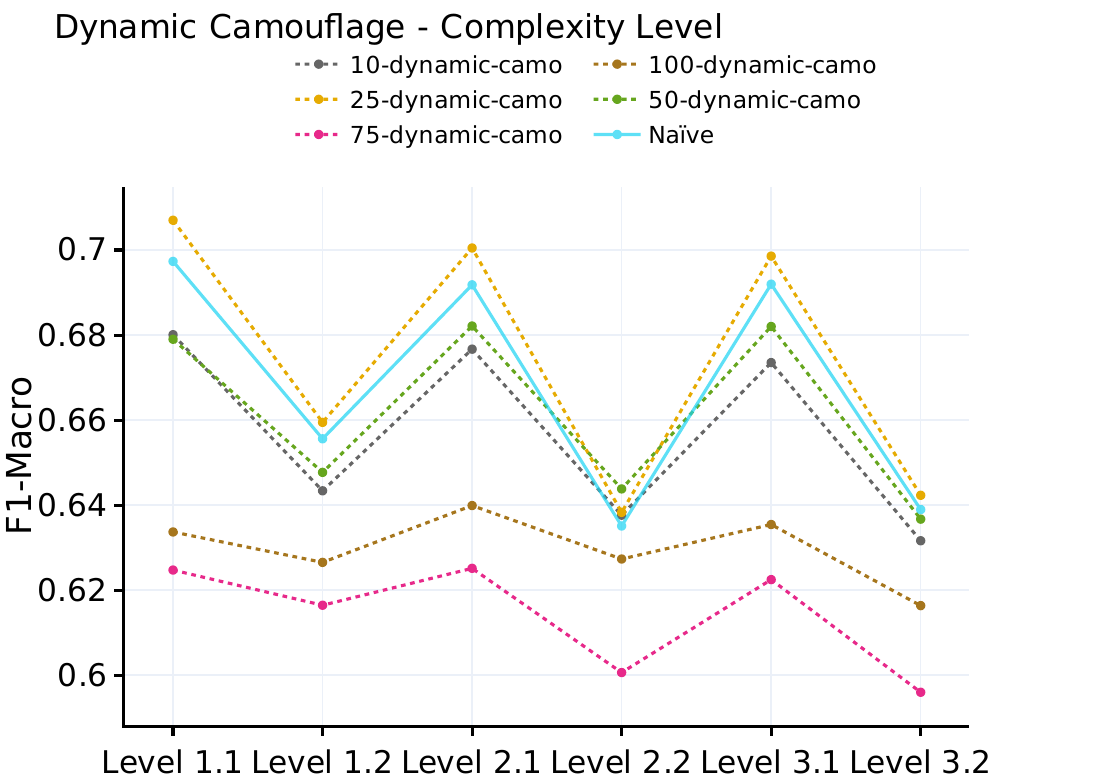}
\caption{}
\label{fig:mbart_offen_var_levels}
\end{subfigure}

\vspace{0.3cm}

\begin{subfigure}[h]{0.48\linewidth}
\includegraphics[width=\linewidth]{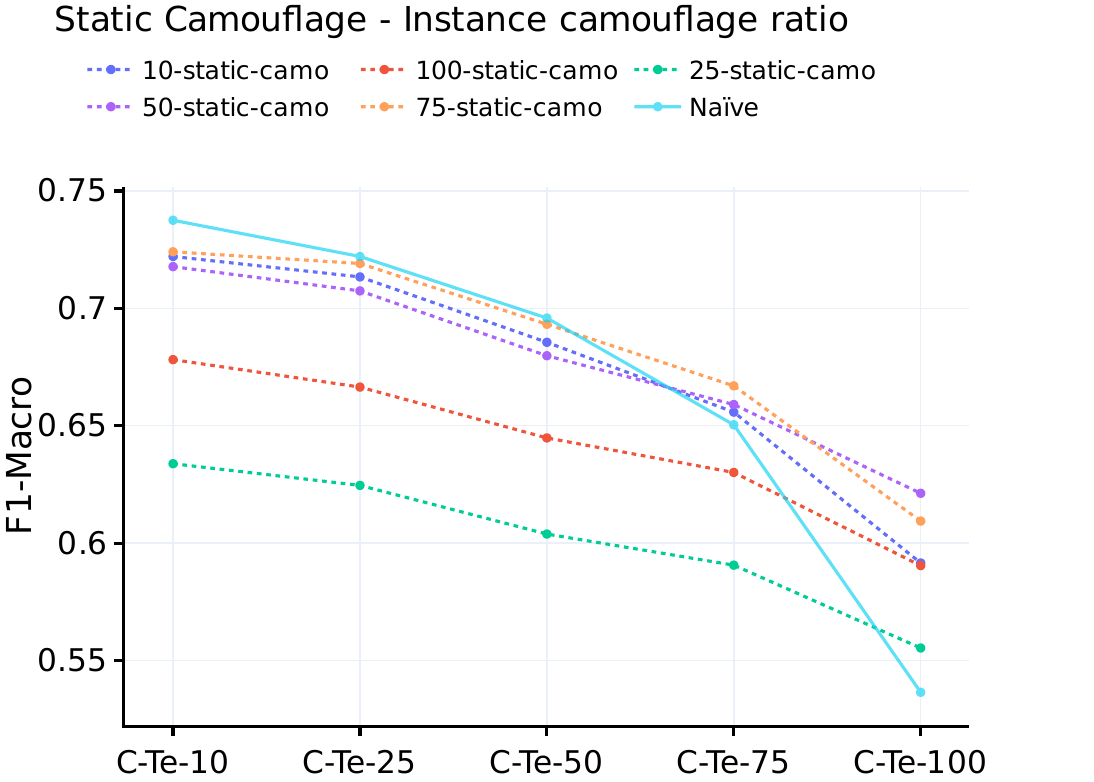}
\caption{}
\label{fig:mbart_offen_pre_percentage}
\end{subfigure}
\begin{subfigure}[h]{0.48\linewidth}
\includegraphics[width=\linewidth]{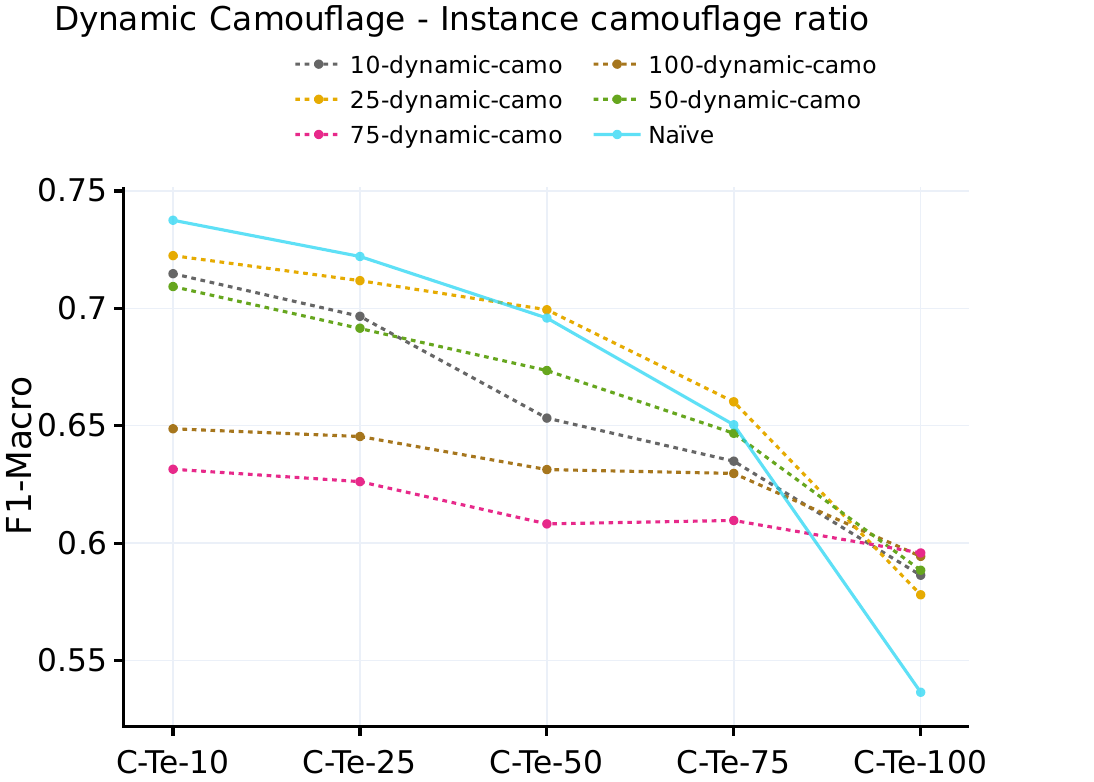}
\caption{}
\label{fig:mbart_offen_var_percentage}
\end{subfigure}

\caption{Comprehensive performance comparison of fine-tuned Encoder-Decoder models against naive models in the Offensive Language task from OffensEval under various conditions. (a) Performance of Pre-camouflaged Models across different levels. (b) Performance of Var-camouflaged Models across different levels. (c) Performance of Pre-camouflaged Models across different camouflage percentages. (d) Performance of Var-camouflaged Models across different camouflage percentages.}
\label{fig:mbart_offen_plots}
\end{figure}

\begin{figure}[tpb]
\textbf{Encoder-Decoder - Constraint Results}
\vspace{0.3cm}
\centering

\begin{subfigure}[h]{0.48\linewidth}
\includegraphics[width=\linewidth]{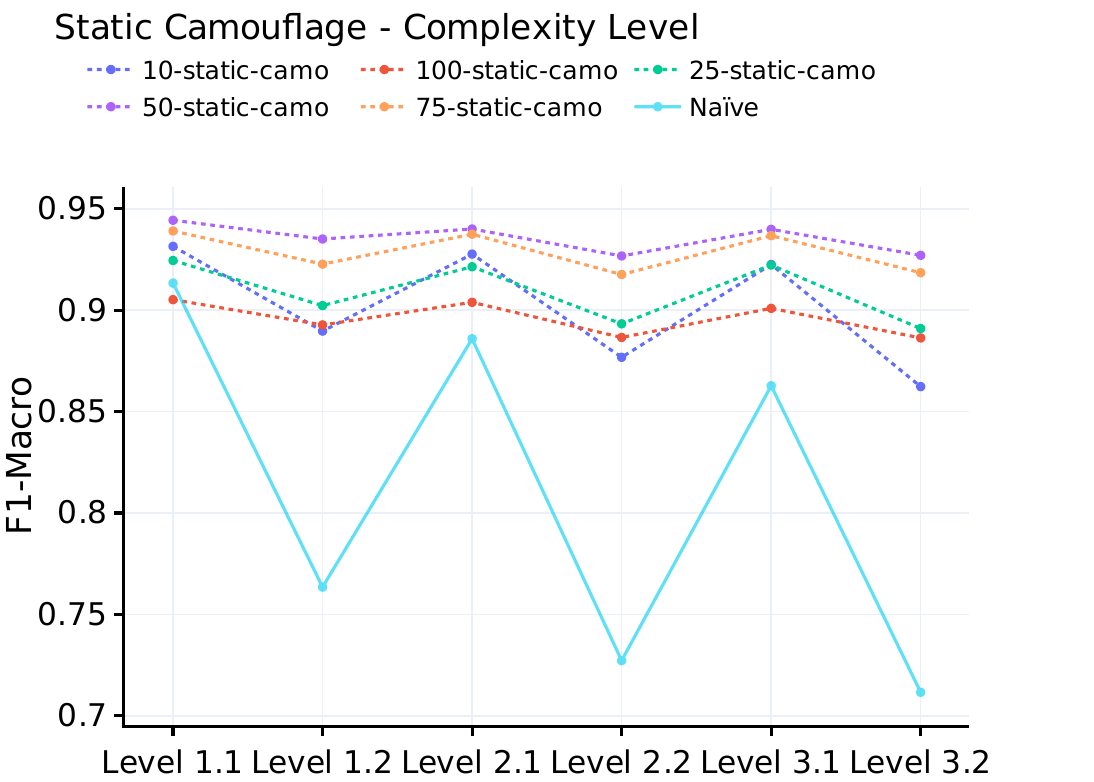}
\caption{}
\label{fig:mbart_constraint_per_levels}
\end{subfigure}
\begin{subfigure}[h]{0.48\linewidth}
\includegraphics[width=\linewidth]{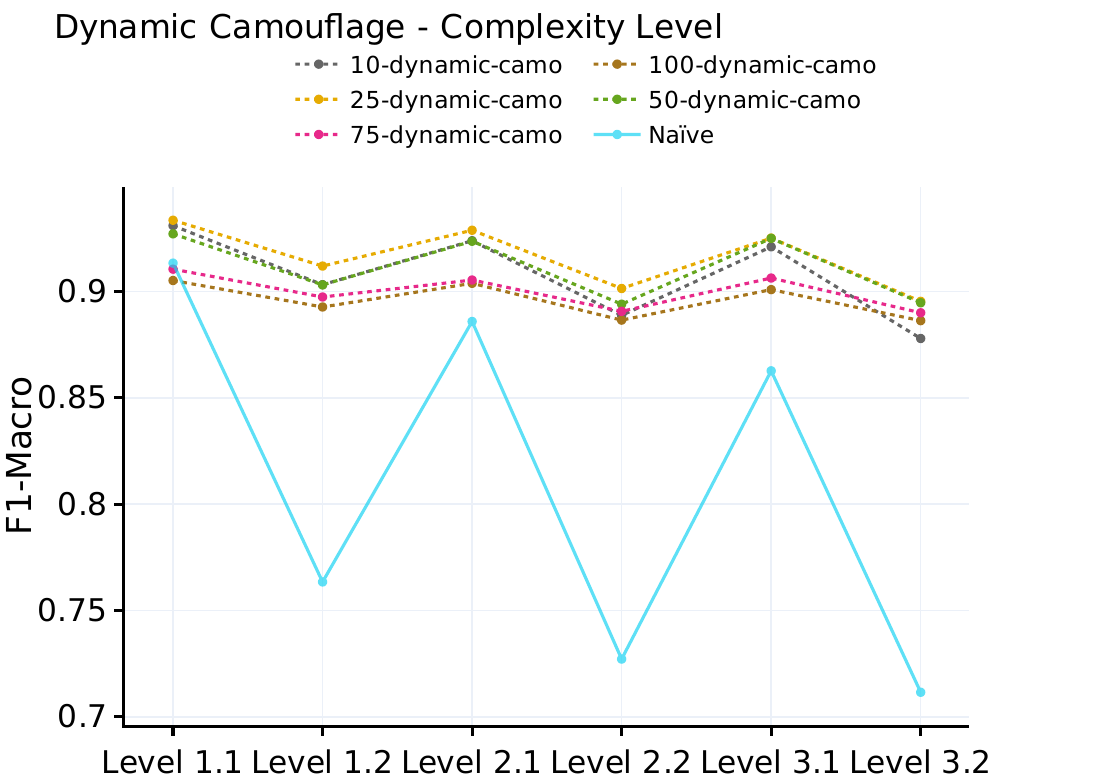}
\caption{}
\label{fig:mbart_constraint_var_levels}
\end{subfigure}

\vspace{0.3cm}

\begin{subfigure}[h]{0.48\linewidth}
\includegraphics[width=\linewidth]{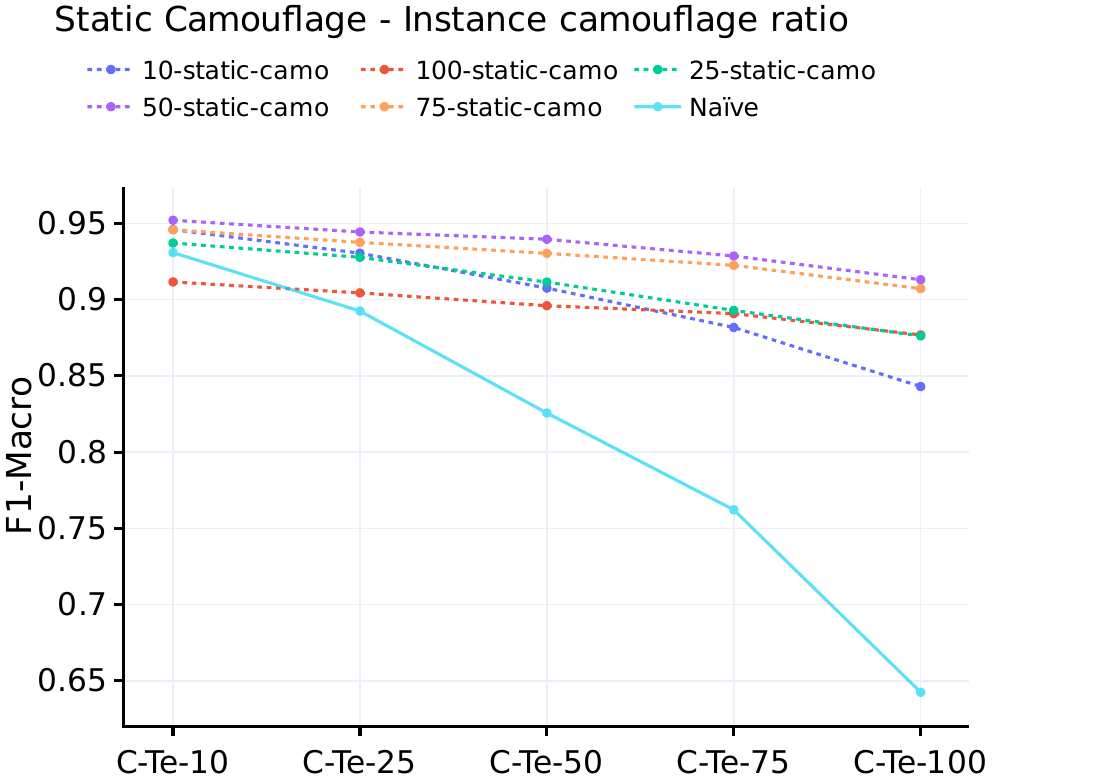}
\caption{}
\label{fig:mbart_constraint_pre_percentage}
\end{subfigure}
\begin{subfigure}[h]{0.48\linewidth}
\includegraphics[width=\linewidth]{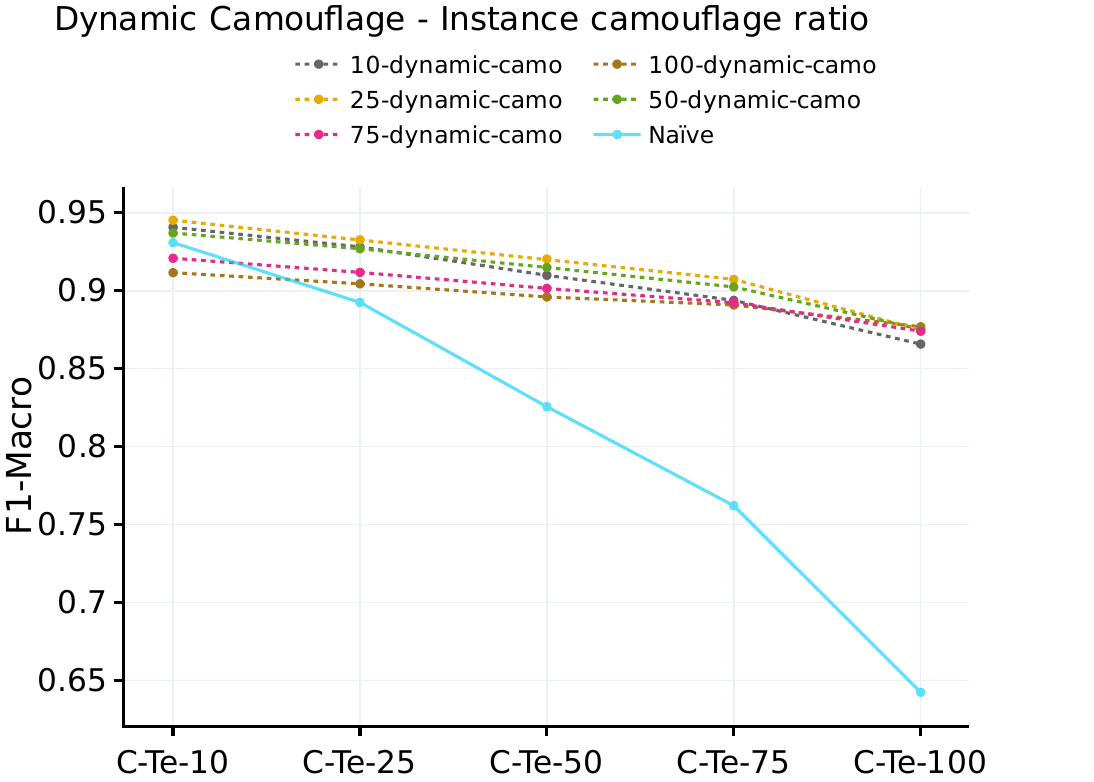}
\caption{}
\label{fig:mbart_constraint_var_percentage}
\end{subfigure}

\caption{Comprehensive performance comparison of fine-tuned Encoder-Decoder models against naive models in the False Information Language task from Constraint under various conditions. (a) Performance of Pre-camouflaged Models across different levels. (b) Performance of Var-camouflaged Models across different levels. (c) Performance of Pre-camouflaged Models across different camouflage percentages. (d) Performance of Var-camouflaged Models across different camouflage percentages.}
\label{fig:mbart_constraint_plots}
\end{figure}

Further evidence of this vulnerability is witnessed in the sharp decline in performance of the Naive model in the face of increasing percentages of camouflaged data. This is clearly depicted in the line plots corresponding to different Transformer configurations (Figures \ref{fig:bert_offen_plot}, \ref{fig:constraint_plots} \ref{fig:pythia_offen_plots}, \ref{fig:pythia_constraint_plots}, \ref{fig:mbart_offen_plots} and \ref{fig:mbart_constraint_plots}). These figures, representing model performance against varying percentages of camouflaged data, mimic real-world scenarios. By simulating circumstances wherein users might employ a spectrum of complexity techniques in word camouflage, these plots offer insightful revelations into how the widespread implementation of such evasion techniques could influence model performance. They suggest potential shifts in performance as word camouflage techniques become more prevalent.

Figures \ref{fig:bert_offen_plot} and \ref{fig:constraint_plots} illustrate the Encoder-only Naive model's performance deterioration as the proportion of camouflaged data instances elevates within both OffensEval and Constraint tasks. A similar trend is discernible in the Decoder-only configuration (Figures \ref{fig:pythia_offen_plots} and \ref{fig:pythia_constraint_plots}) and in the Encoder-Decoder configuration (Figures \ref{fig:mbart_offen_plots} and \ref{fig:mbart_constraint_plots}).

In the following section, it will be demonstrated that this performance decline is more pronounced in Naive models compared to their adversarially-trained counterparts, further underscoring the susceptibility of Naive models to adversarial attacks.

Overall, these results bring to light the inherent susceptibility of Naive models across all transformer configurations to word camouflage attacks, thereby emphasizing the necessity for solutions aimed at fortifying these vulnerabilities. 


\subsection{Phase II: Enhancing Transformer Robustness Against Word Camouflage Attacks}

The motivation behind the second phase of the study was to enhance the robustness of Transformer models against word camouflage attacks. In the first phase, it was determined that word camouflage attacks could significantly degrade the performance of the models. This phase aimed to evaluate the effectiveness of various countermeasures, mainly through the integration of adversarial training strategies using different proportions and methods of data camouflage.

\subsubsection{Static vs Dynamic Camouflage}

The analysis of the experiments revealed that models trained with a mix of original and statically camouflaged data, up to a 75\% proportion, performed admirably across a variety language detection tasks in both the OffensEval and Constraint contexts. Specifically, in the OffensEval setting, the Encoder-only model trained with 75\% statically camouflaged data demonstrated improving the  performance and decreasing the performance reduction with complexity (see \ref{table:bert_weakness_offen}). However, models trained on completely statically camouflaged datasets saw a significant drop in performance, with the F1-Macro score falling to 0.4189 across all levels for the `100-static' Encoder-only model (see \ref{table:bert_weakness_offen}) and 0.3436 for the `100-dynamic' Decoder-only model (see \ref{table:pythia_weakness_constraint}). This degradation in performance supports the conclusion that a balanced mix of original and camouflaged data produces superior results.

On the other hand, dynamic camouflage strategies displayed a distinct advantage, with models incorporating these techniques showing substantial robustness. For instance, a model that dynamically camouflaged 100\% of the data during training managed to avoid performance deterioration and exhibited marked robustness in Table \ref{table:bert_weakness_offen} while the 100\% static counterpart stuck. This observation suggests that dynamic camouflage introduces a certain degree of variability and richness into the training data, which in turn enables the model to learn more effectively and flexibly, although it is still recommended to include some percentage of original, non-camouflaged data in the training process.

\subsubsection{Effect of Camouflage Complexity}

An intriguing trend was revealed when examining the relationship between the complexity of camouflage techniques and model performance. As the camouflage complexity level increased, naive models exhibited more pronounced performance reduction, as previously observed in Section~\ref{sec:phase1} in both the OffensEval and Constraint tasks. This trend implies an inherent vulnerability of Naive models to heightened complexity in adversarial attacks. However, adversarially trained models, particularly those trained with a combination of original and dynamic camouflage data, such as the `75-dynamic-camo' model, demonstrated lower degrees of reduction (Tables \ref{table:bert_weakness}, \ref{table:pythia_weakness} and \ref{table:mbart_weakness}), thereby highlighting the resilience of adversarially trained models and their enhanced ability to counter advanced camouflage attacks.

\subsubsection{Influence of Camouflage Percentage}

The analysis also brought to light an inverse relationship between the percentage of camouflaged data and model performance. As the proportion of camouflaged data in the test set increased, the model's performance correspondingly declined. The trend was evident across all models, indicating the importance of considering the expected degree of camouflage in the actual scenario in order to estimate the impact and to select the most appropriate adversarial training method.

\subsubsection{Architectural Considerations}

In addition to the above, an underlying trend emerged from the analysis, which transcended the specific architectural configurations: training with dynamic camouflage consistently demonstrated superior resilience to adversarial attacks. This strategy maintained or even enhanced the original model's performance in many cases, while a high camouflage percentage brought significant challenges.

When assessing the impact of different model architectures, it was observed that Encoder-only models demonstrated significant robustness against adversarial attacks when trained with both static and dynamic camouflage techniques. In the Constraint task, for example, the `75-static-camo' and `25-static-camo' Encoder-only models outperformed their counterparts on several levels. On the other hand, Encoder-Decoder and Decoder-only setups also showcased improved performance, thus affirming that adversarial training strategies can be effectively utilized across diverse architectural setups.


Finally, a key finding was the inverse relationship between the proportion of camouflaged data in the test set and the model's performance. As the proportion of camouflaged data increased, there was a noticeable decline in performance across all models. Thus, striking a balance in data camouflaging is vital: while it can boost generalization, excessive usage may compromise performance.

\section{Conclusion}

In response to the escalating prominence of adversarial attacks in Natural Language Processing (NLP), this research presented a two-phase methodology to enhance Transformer models' robustness. In the first phase, the susceptibility of naive models to camouflage adversarial attacks was identified, demonstrating a clear need for improved defences.

A proactive strategy, incorporating adversarial training with both static and dynamic camouflage, was introduced in the second phase. The models trained with a small proportion (10-25\%) of statically camouflaged data outperformed those trained entirely with static camouflage. Dynamic camouflage introduced during training further boosted the models' learning and generalization capabilities.

In the comparison of various configurations, encoder-only models often excelled in managing adversarial attacks, underscoring their superior adaptability. However, all configurations faced difficulties as the proportion of camouflaged data increased, which emphasized the importance of balancing between original and camouflaged data in the training set.

These findings carry significant implications for improving AI system robustness. Nevertheless, the limitations of this study are acknowledged. The proposed approach's effectiveness may fluctuate based on camouflage complexity and the type of data encountered. Moreover, this research primarily concentrated on black-box adversarial attacks, leaving other types largely unexplored.

Future research could expand this methodology to other adversarial attack types and model architectures, and further explore the influence of camouflage complexity and type on model learning and robustness.



\bibliographystyle{elsarticle-num}
\bibliography{references}

\end{document}

%% file: table_parameters.tex
\caption{Parameters defining the levels of word camouflage complexity considered. The table displays three levels of complexity, with each level having two versions based on the max\_top\_n parameter set to either 5 or 20 for versions 1 and 2, respectively. This parameter defines the maximum number of keywords to extract and camouflage. These levels and versions illustrate the diverse configurations for evaluating the robustness of language models against various camouflage techniques.}
\label{table_parameters}
\resizebox{\columnwidth}{!}{%
\begin{tblr}{
  cell{1}{2} = {c},
  hlines,
  hline{1,5} = {-}{0.08em},
}
        & \textbf{Parameters}                                                                                                                                                                                                                                                                                                           \\
Level 1 & {max\_top\_n=[5, 20]\\leet\_punt\_prb=0.9\\leet\_change\_prb=0.8\\leet\_change\_frq=0.8\\leet\_uniform\_change=0.5\\method=["basic\_leetspeak"]}                                                                                                                                                                              \\
Level 2 & {max\_top\_n=[5,20]\\ leet\_punt\_prb=0.9\\ leet\_change\_prb=0.5\\ leet\_change\_frq=0.8\\ leet\_uniform\_change=0.6\\ punt\_hyphenate\_prb=0.7\\ punt\_uniform\_change\_prb=0.95\\ punt\_word\_splitting\_prb=0.8\\method=[ïntermediate\_leetspeak "punct\_camo"]}                                                          \\
Level 3 & {max\_top\_n=[5,20]\\ leet\_punt\_prb=0.4\\ leet\_change\_prb=0.5\\ leet\_change\_frq=0.8\\ leet\_uniform\_change=0.6\\ punt\_hyphenate\_prb=0.7\\ punt\_uniform\_change\_prb=0.95\\ punt\_word\_splitting\_prb=0.8\\ inv\_max\_dist=4\\ inv\_only\_max\_dist\_prb=0.5\\method=[ädvanced\_leetspeak "punct\_camo ïnv\_camo"]} 
\end{tblr}
}

%% file: table_training_parameters.tex
\centering
\captionof{table}{ 
 A description of the parameters considered during the training of the models for word camouflaged Named Entity Recognition with Spacy.}\label{table:training-params}
\resizebox{\columnwidth}{!}{%
\begin{tabular}{cl} \toprule
\multirow{4}{*}{learning rate}                          & \textcolor[rgb]{0.102,0.11,0.122}{initial\_rate =~}\textcolor[rgb]{0.102,0.11,0.122}{0.00005}                                  \\
                                                        & \textcolor[rgb]{0.102,0.11,0.122}{total\_steps =~}\textcolor[rgb]{0.102,0.11,0.122}{20000}                                     \\
                                                        & scheduler =~\textcolor[rgb]{0.102,0.11,0.122}{warmup\_linear}                                                                  \\
                                                        & \textcolor[rgb]{0.212,0.227,0.239}{\textcolor[rgb]{0.102,0.11,0.122}{warmup\_steps =~}\textcolor[rgb]{0.102,0.11,0.122}{250}}  \\ \hline
\multirow{3}{*}{epochs}                                 & \textcolor[rgb]{0.212,0.227,0.239}{\textcolor[rgb]{0.102,0.11,0.122}{max\_epochs =~}\textcolor[rgb]{0.102,0.11,0.122}{0}}      \\
                                                        & \textcolor[rgb]{0.102,0.11,0.122}{max\_steps =~}\textcolor[rgb]{0.102,0.11,0.122}{20000}                                       \\
                                                        & \textcolor[rgb]{0.102,0.11,0.122}{patience =~}\textcolor[rgb]{0.102,0.11,0.122}{1600}                                          \\ \hline
\textcolor[rgb]{0.102,0.11,0.122}{accumulate\_gradient} & 3                                                                                                                              \\ \hline
\multirow{7}{*}{optimizer}                              & \textcolor[rgb]{0.212,0.227,0.239}{\textcolor[rgb]{0.102,0.11,0.122}{AdamW}}                                                   \\
                                                        & \textcolor[rgb]{0.102,0.11,0.122}{beta = 1}\textcolor[rgb]{0.102,0.11,0.122}{0.9}                                              \\
                                                        & \textcolor[rgb]{0.102,0.11,0.122}{beta2 =~}\textcolor[rgb]{0.102,0.11,0.122}{0.999}                                            \\
                                                        & \textcolor[rgb]{0.102,0.11,0.122}{eps =~}\textcolor[rgb]{0.102,0.11,0.122}{1e-8}                                               \\
                                                        & \textcolor[rgb]{0.102,0.11,0.122}{grad\_clip =~}\textcolor[rgb]{0.102,0.11,0.122}{1}                                           \\
                                                        & \textcolor[rgb]{0.102,0.11,0.122}{l2 =~}\textcolor[rgb]{0.102,0.11,0.122}{0.01}                                                \\
                                                        & \textcolor[rgb]{0.102,0.11,0.122}{l2\_is\_weight\_decay =~}\textcolor[rgb]{0.102,0.11,0.122}{true}                             \\ \hline
eval\_frequency                                         & 200                                                                                                                            \\ \hline
dropout                                                 & 0.1                                                                                                                            \\ \bottomrule
\end{tabular}
}

%% file: table_bert_offen_weakness.tex
\centering
\resizebox{\linewidth}{!}{%
\begin{tblr}{
  row{1} = {c},
  row{3} = {c},
  cell{1}{1} = {r=3}{},
  cell{1}{2} = {r=3}{},
  cell{1}{3} = {c=8}{},
  cell{2}{3} = {c=7}{c},
  cell{2}{10} = {r=2}{},
  cell{4}{2} = {c},
  cell{4}{3} = {c},
  cell{4}{4} = {c},
  cell{4}{5} = {c},
  cell{4}{6} = {c},
  cell{4}{7} = {c},
  cell{4}{8} = {c},
  cell{4}{9} = {c},
  cell{4}{10} = {c},
  cell{5}{2} = {c},
  cell{5}{3} = {c},
  cell{5}{4} = {c},
  cell{5}{5} = {c},
  cell{5}{6} = {c},
  cell{5}{7} = {c},
  cell{5}{8} = {c},
  cell{5}{9} = {c},
  cell{5}{10} = {c},
  cell{6}{2} = {c},
  cell{6}{3} = {c},
  cell{6}{4} = {c},
  cell{6}{5} = {c},
  cell{6}{6} = {c},
  cell{6}{7} = {c},
  cell{6}{8} = {c},
  cell{6}{9} = {c},
  cell{6}{10} = {c},
  cell{7}{2} = {c},
  cell{7}{3} = {c},
  cell{7}{4} = {c},
  cell{7}{5} = {c},
  cell{7}{6} = {c},
  cell{7}{7} = {c},
  cell{7}{8} = {c},
  cell{7}{9} = {c},
  cell{7}{10} = {c},
  cell{8}{2} = {c},
  cell{8}{3} = {c},
  cell{8}{4} = {c},
  cell{8}{5} = {c},
  cell{8}{6} = {c},
  cell{8}{7} = {c},
  cell{8}{8} = {c},
  cell{8}{9} = {c},
  cell{8}{10} = {c},
  cell{9}{2} = {c},
  cell{9}{3} = {c},
  cell{9}{4} = {c},
  cell{9}{5} = {c},
  cell{9}{6} = {c},
  cell{9}{7} = {c},
  cell{9}{8} = {c},
  cell{9}{9} = {c},
  cell{9}{10} = {c},
  cell{10}{2} = {c},
  cell{10}{3} = {c},
  cell{10}{4} = {c},
  cell{10}{5} = {c},
  cell{10}{6} = {c},
  cell{10}{7} = {c},
  cell{10}{8} = {c},
  cell{10}{9} = {c},
  cell{10}{10} = {c},
  cell{11}{2} = {c},
  cell{11}{3} = {c},
  cell{11}{4} = {c},
  cell{11}{5} = {c},
  cell{11}{6} = {c},
  cell{11}{7} = {c},
  cell{11}{8} = {c},
  cell{11}{9} = {c},
  cell{11}{10} = {c},
  cell{12}{2} = {c},
  cell{12}{3} = {c},
  cell{12}{4} = {c},
  cell{12}{5} = {c},
  cell{12}{6} = {c},
  cell{12}{7} = {c},
  cell{12}{8} = {c},
  cell{12}{9} = {c},
  cell{12}{10} = {c},
  cell{13}{2} = {c},
  cell{13}{3} = {c},
  cell{13}{4} = {c},
  cell{13}{5} = {c},
  cell{13}{6} = {c},
  cell{13}{7} = {c},
  cell{13}{8} = {c},
  cell{13}{9} = {c},
  cell{13}{10} = {c},
  cell{14}{2} = {c},
  cell{14}{3} = {c},
  cell{14}{4} = {c},
  cell{14}{5} = {c},
  cell{14}{6} = {c},
  cell{14}{7} = {c},
  cell{14}{8} = {c},
  cell{14}{9} = {c},
  cell{14}{10} = {c},
  vline{2-3} = {4-14}{},
  hline{1,4-5,10,15} = {-}{},
  hline{3} = {3-9}{},
}
\textbf{Model } & \textbf{F1-Macro } & \textbf{Performance Reduction} &              &              &              &              &               &                       &                \\
                &                    & \textbf{Levels}                         &              &              &              &              &               &                       & \textbf{AugLy} \\
                &                    & \textbf{1.1}                   & \textbf{1.2} & \textbf{2.1} & \textbf{2.2} & \textbf{3.1} & \textbf{3.2}  & Avg                   &                \\
Naïve           & 0.7782             & 2\%                            & 6\%          & 6\%          & 13\%         & 7\%          & 14\%          & 8\%                   & 10\%           \\
10-static       & 0.7870*            & 2\%                            & \textbf{5\%} & 6\%          & 11\%         & 6\%          & 13\%          & 7\%                   & 9\%            \\
25-static       & 0.7852*            & 2\%                            & \textbf{5\%} & 6\%          & 10\%         & \textbf{5\%} & 12\%          & \textbf{\textbf{6\%}} & 9\%            \\
50-static       & 0.7887*            & 2\%                            & 7\%          & \textbf{5\%} & 11\%         & \textbf{5\%} & 11\%          & 7\%                   & \textbf{8\%}   \\
75-static       & 0.7962*            & 3\%                            & 7\%          & \textbf{5\%} & \textbf{9\%} & \textbf{5\%} & \textbf{10\%} & 7\%                   & \textbf{8\%}   \\
100-static      & 0.4189             & -                              & -            & -            & -            & -            & -             & -                     & -              \\
10-dynamic      & 0.7828*            & 3\%                            & 7\%          & 5\%          & 11\%         & \textbf{5\%} & 13\%          & 7\%                   & 10\%           \\
25-dynamic      & 0.7947*            & 4\%                            & 8\%          & 7\%          & 12\%         & 7\%          & 13\%          & 8\%                   & 10\%           \\
50-dynamic      & 0.7791             & 3\%                            & 8\%          & 5\%          & 9\%          & 6\%          & \textbf{9\%}  & 7\%                   & 8\%            \\
75-dynamic      & 0.7945*            & 3\%                            & 6\%          & 5\%          & 9\%          & 6\%          & \textbf{9\%}  & 6\%                   & 7\%            \\
100-dynamic     & 0.7527             & \textbf{0\%}                   & \textbf{4\%} & \textbf{3\%} & \textbf{8\%} & \textbf{4\%} & 10\%          & \textbf{\textbf{5\%}} & \textbf{5\%}   
\end{tblr}
}

%% file: table_bert_constraint_weakness.tex

\centering
\resizebox{\linewidth}{!}{%
\centering
\begin{tblr}{
  row{1} = {c},
  column{4} = {c},
  column{5} = {c},
  column{6} = {c},
  column{7} = {c},
  column{8} = {c},
  cell{1}{1} = {r=3}{},
  cell{1}{2} = {r=3}{},
  cell{1}{3} = {c=8}{},
  cell{2}{3} = {c=7}{c},
  cell{2}{10} = {r=2}{},
  cell{3}{3} = {c},
  cell{4}{2} = {c},
  cell{4}{3} = {c},
  cell{4}{10} = {c},
  cell{5}{2} = {c},
  cell{5}{3} = {c},
  cell{5}{10} = {c},
  cell{6}{2} = {c},
  cell{6}{3} = {c},
  cell{6}{10} = {c},
  cell{7}{2} = {c},
  cell{7}{3} = {c},
  cell{7}{10} = {c},
  cell{8}{2} = {c},
  cell{8}{3} = {c},
  cell{8}{10} = {c},
  cell{9}{2} = {c},
  cell{9}{3} = {c},
  cell{9}{10} = {c},
  cell{10}{2} = {c},
  cell{10}{3} = {c},
  cell{10}{10} = {c},
  cell{11}{2} = {c},
  cell{11}{3} = {c},
  cell{11}{10} = {c},
  cell{12}{2} = {c},
  cell{12}{3} = {c},
  cell{12}{10} = {c},
  cell{13}{2} = {c},
  cell{13}{3} = {c},
  cell{13}{10} = {c},
  cell{14}{2} = {c},
  cell{14}{3} = {c},
  cell{14}{10} = {c},
  vline{2-3,10} = {4-14}{},
  hline{1,4-5,10,15} = {-}{},
  hline{3} = {3-9}{},
}
\textbf{Model } & \textbf{F1-Macro } & \textbf{Performance Reduction} &              &              &              &              &               &                       &                \\
                &                    & \textbf{Levels}                         &              &              &              &              &               &                       & \textbf{AugLy} \\
                &                    & \textbf{1.1}                   & \textbf{1.2} & \textbf{2.1} & \textbf{2.2} & \textbf{3.1} & \textbf{3.2}  & Avg                   &           \\
Naïve       & 0.9677   & 4\%                   & 14\%         & 7\%          & 17\%         & 11\%         & 21\%         & 12\%                  & 9\%          \\
10-static   & 0.9649   & \textbf{1\%}          & 3\%          & \textbf{1\%} & 6\%          & 3\%          & 6\%          & 3\%                   & 7\%          \\
25-static   & 0.9649   & \textbf{1\%}          & 3\%          & 2\%          & 5\%          & 2\%          & 5\%          & 3\%                   & 6\%          \\
50-static   & 0.9602   & \textbf{1\%}          & 3\%          & 2\%          & 4\%          & 2\%          & 4\%          & 3\%                   & \textbf{4\%} \\
75-static   & 0.9517   & \textbf{1\%}          & 2\%          & \textbf{1\%} & 3\%          & 2\%          & 4\%          & \textbf{\textbf{2\%}} & 7\%          \\
100-static  & 0.9443   & \textbf{1\%}          & \textbf{1\%} & \textbf{1\%} & 3\%          & \textbf{1\%} & \textbf{3\%} & \textbf{\textbf{2\%}} & 8\%          \\
10-dynamic  & 0.9629   & 2\%                   & 3\%          & 3\%          & 5\%          & 3\%          & 6\%          & 4\%                   & 6\%          \\
25-dynamic  & 0.9653   & 2\%                   & 3\%          & \textbf{1\%} & 5\%          & 3\%          & 5\%          & 3\%                   & 6\%          \\
50-dynamic  & 0.9569   & \textbf{1\%}          & 2\%          & 2\%          & 4\%          & 2\%          & 5\%          & 3\%                   & 7\%          \\
75-dynamic  & 0.9569   & \textbf{1\%}          & 2\%          & 2\%          & 3\%          & 2\%          & 4\%          & \textbf{\textbf{2\%}} & 6\%          \\
100-dynamic & 0.9427   & \textbf{1\%}          & \textbf{1\%} & \textbf{1\%} & \textbf{2\%} & \textbf{1\%} & \textbf{3\%} & \textbf{\textbf{2\%}} & \textbf{5\%} 
\end{tblr}
}

%% file: table_pythia_offen_weakness.tex
\centering
\resizebox{\linewidth}{!}{%
\begin{tblr}{
  row{1} = {c},
  column{4} = {c},
  column{5} = {c},
  column{6} = {c},
  column{7} = {c},
  column{8} = {c},
  cell{1}{1} = {r=3}{},
  cell{1}{2} = {r=3}{},
  cell{1}{3} = {c=8}{},
  cell{2}{3} = {c=7}{c},
  cell{2}{10} = {r=2}{},
  cell{3}{3} = {c},
  cell{4}{2} = {c},
  cell{4}{3} = {c},
  cell{4}{10} = {c},
  cell{5}{2} = {c},
  cell{5}{3} = {c},
  cell{5}{10} = {c},
  cell{6}{2} = {c},
  cell{6}{3} = {c},
  cell{6}{10} = {c},
  cell{7}{2} = {c},
  cell{7}{3} = {c},
  cell{7}{10} = {c},
  cell{8}{2} = {c},
  cell{8}{3} = {c},
  cell{8}{10} = {c},
  cell{9}{2} = {c},
  cell{9}{3} = {c},
  cell{9}{10} = {c},
  cell{10}{2} = {c},
  cell{10}{3} = {c},
  cell{10}{10} = {c},
  cell{11}{2} = {c},
  cell{11}{3} = {c},
  cell{11}{10} = {c},
  cell{12}{2} = {c},
  cell{12}{3} = {c},
  cell{12}{10} = {c},
  cell{13}{2} = {c},
  cell{13}{3} = {c},
  cell{13}{10} = {c},
  cell{14}{2} = {c},
  cell{14}{3} = {c},
  cell{14}{10} = {c},
  vline{2-3} = {4-14}{},
  hline{1,4-5,10,15} = {-}{},
  hline{3} = {3-9}{},
}
\textbf{Model}   & \textbf{F1-Macro } & \textbf{Performance Reduction} &               &              &              &              &              &                       &                \\
                 &                    & \textbf{Levels}                &               &              &              &              &              &                       & \textbf{AugLy} \\
                 &                    & \textbf{1.1}                   & \textbf{~1.2} & \textbf{2.1} & \textbf{2.2} & \textbf{3.1} & \textbf{3.2} & \textbf{Avg}          &                \\
Naïve            & 0.7185             & 7\%                            & 15\%          & 8\%          & 16\%         & 8\%          & 16\%         & 12\%                  & 11\%           \\
10-static   & 0.7159             & 8\%                            & 12\%          & 9\%          & 13\%         & 8\%          & 14\%         & 11\%                  & 14\%           \\
25-static   & 0.6906             & 4\%                            & 11\%          & 5\%          & 12\%         & 5\%          & 11\%         & 8\%                   & 8\%            \\
50-static   & 0.6750             & 4\%                            & 9\%           & 5\%          & 10\%         & 6\%          & 10\%         & 7\%                   & 7\%            \\
75-static   & 0.6249             & 7\%                            & 9\%           & 5\%          & 7\%          & 4\%          & 8\%          & 7\%                   & 8\%            \\
100-static  & 0.4309             & \textbf{1\%}                   & \textbf{1\%}  & \textbf{1\%} & \textbf{1\%} & \textbf{1\%} & \textbf{1\%} & \textbf{\textbf{1\%}} & \textbf{0\%}   \\
10-dynamic  & 0.7351*            & 6\%                            & 13\%          & 6\%          & 15\%         & 7\%          & 15\%         & 10\%                  & 9\%            \\
25-dynamic  & 0.7368*            & 5\%                            & 12\%          & 7\%          & 17\%         & 9\%          & 13\%         & 11\%                  & 10\%           \\
50-dynamic  & 0.6636             & 3\%                            & 9\%           & 4\%          & 10\%         & 5\%          & 10\%         & 7\%                   & 4\%            \\
75-dynamic  & 0.7029             & 7\%                            & 11\%          & 8\%          & 11\%         & 7\%          & 11\%         & 9\%                   & 10\%           \\
100-dynamic & 0.5932             & \textbf{2\%}                   & \textbf{4\%}  & \textbf{3\%} & \textbf{8\%} & \textbf{3\%} & \textbf{6\%} & \textbf{\textbf{5\%}} & \textbf{5\%}   
\end{tblr}
}

%% file: table_pythia_constraint_weakness.tex
\centering
\resizebox{\linewidth}{!}{%
\begin{tblr}{
  row{1} = {c},
  row{3} = {c},
  cell{1}{1} = {r=3}{},
  cell{1}{2} = {r=3}{},
  cell{1}{3} = {c=8}{},
  cell{2}{3} = {c=7}{c},
  cell{4}{2} = {c},
  cell{4}{3} = {c},
  cell{4}{4} = {c},
  cell{4}{5} = {c},
  cell{4}{6} = {c},
  cell{4}{7} = {c},
  cell{4}{8} = {c},
  cell{4}{9} = {c},
  cell{4}{10} = {c},
  cell{5}{2} = {c},
  cell{5}{3} = {c},
  cell{5}{4} = {c},
  cell{5}{5} = {c},
  cell{5}{6} = {c},
  cell{5}{7} = {c},
  cell{5}{8} = {c},
  cell{5}{9} = {c},
  cell{5}{10} = {c},
  cell{6}{2} = {c},
  cell{6}{3} = {c},
  cell{6}{4} = {c},
  cell{6}{5} = {c},
  cell{6}{6} = {c},
  cell{6}{7} = {c},
  cell{6}{8} = {c},
  cell{6}{9} = {c},
  cell{6}{10} = {c},
  cell{7}{2} = {c},
  cell{7}{3} = {c},
  cell{7}{4} = {c},
  cell{7}{5} = {c},
  cell{7}{6} = {c},
  cell{7}{7} = {c},
  cell{7}{8} = {c},
  cell{7}{9} = {c},
  cell{7}{10} = {c},
  cell{8}{2} = {c},
  cell{8}{3} = {c},
  cell{8}{4} = {c},
  cell{8}{5} = {c},
  cell{8}{6} = {c},
  cell{8}{7} = {c},
  cell{8}{8} = {c},
  cell{8}{9} = {c},
  cell{8}{10} = {c},
  cell{9}{2} = {c},
  cell{9}{3} = {c},
  cell{9}{4} = {c},
  cell{9}{5} = {c},
  cell{9}{6} = {c},
  cell{9}{7} = {c},
  cell{9}{8} = {c},
  cell{9}{9} = {c},
  cell{9}{10} = {c},
  cell{10}{2} = {c},
  cell{10}{3} = {c},
  cell{10}{4} = {c},
  cell{10}{5} = {c},
  cell{10}{6} = {c},
  cell{10}{7} = {c},
  cell{10}{8} = {c},
  cell{10}{9} = {c},
  cell{10}{10} = {c},
  cell{11}{2} = {c},
  cell{11}{3} = {c},
  cell{11}{4} = {c},
  cell{11}{5} = {c},
  cell{11}{6} = {c},
  cell{11}{7} = {c},
  cell{11}{8} = {c},
  cell{11}{9} = {c},
  cell{11}{10} = {c},
  cell{12}{2} = {c},
  cell{12}{3} = {c},
  cell{12}{4} = {c},
  cell{12}{5} = {c},
  cell{12}{6} = {c},
  cell{12}{7} = {c},
  cell{12}{8} = {c},
  cell{12}{9} = {c},
  cell{12}{10} = {c},
  cell{13}{2} = {c},
  cell{13}{3} = {c},
  cell{13}{4} = {c},
  cell{13}{5} = {c},
  cell{13}{6} = {c},
  cell{13}{7} = {c},
  cell{13}{8} = {c},
  cell{13}{9} = {c},
  cell{13}{10} = {c},
  cell{14}{2} = {c},
  cell{14}{3} = {c},
  cell{14}{4} = {c},
  cell{14}{5} = {c},
  cell{14}{6} = {c},
  cell{14}{7} = {c},
  cell{14}{8} = {c},
  cell{14}{9} = {c},
  cell{14}{10} = {c},
  vline{2-3} = {4-14}{},
  hline{1,4-5,10,15} = {-}{},
  hline{3} = {3-9}{},
}
\textbf{Model } & {\textbf{Test}\\\textbf{F1-Macro }} & \textbf{Performance Reduction} &              &              &               &              &              &                       &                \\
                &                                     & \textbf{Levels}                &              &              &               &              &              &                       &                \\
                &                                     & \textbf{1.1}                   & \textbf{1.2} & \textbf{2.1} & \textbf{~2.2} & \textbf{3.1} & \textbf{3.2} & \textbf{Avg}          & \textbf{AugLy} \\
Naïve           & 0.9380                              & 4\%                            & 10\%         & 8\%          & 15\%          & 7\%          & 16\%         & 10\%                  & 7\%            \\
10-static       & 0.9270                              & 3\%                            & 11\%         & 4\%          & 12\%          & 4\%          & 11\%         & 7\%                   & 8\%            \\
25-static       & 0.9046                              & 3\%                            & 6\%          & 4\%          & 9\%           & 4\%          & 9\%          & 6\%                   & 6\%            \\
50-static       & 0.8975                              & 2\%                            & 5\%          & 2\%          & 6\%           & 3\%          & 7\%          & 4\%                   & 4\%            \\
75-static       & 0.8700                              & 3\%                            & 6\%          & 3\%          & 6\%           & 3\%          & 6\%          & 5\%                   & 5\%            \\
100-static      & 0.7935                              & \textbf{1\%}                   & \textbf{1\%} & \textbf{1\%} & \textbf{1\%}  & \textbf{1\%} & \textbf{2\%} & \textbf{\textbf{1\%}} & \textbf{2\%}   \\
10-dynamic      & 0.9340                              & 3\%                            & 7\%          & 4\%          & 8\%           & 4\%          & 9\%          & 6\%                   & 8\%            \\
25-dynamic      & 0.9198                              & 3\%                            & 9\%          & 4\%          & 12\%          & 6\%          & 11\%         & 7\%                   & 6\%            \\
50-dynamic      & 0.8752                              & 2\%                            & 5\%          & \textbf{2\%} & 6\%           & 3\%          & 6\%          & 4\%                   & \textbf{4\%}   \\
75-dynamic      & 0.8958                              & \textbf{1\%}                   & \textbf{3\%} & \textbf{2\%} & \textbf{3\%}  & \textbf{2\%} & \textbf{4\%} & \textbf{\textbf{2\%}} & \textbf{4\%}   \\
100-dynamic     & 0.3436                              & -                              & -            & -            & -             & -            & -            & -                     & -              
\end{tblr}
}

%% file: table_mbart_offen_weakness.tex
\centering
\resizebox{\linewidth}{!}{%
\begin{tblr}{
  row{1} = {c},
  row{3} = {c},
  cell{1}{1} = {r=3}{},
  cell{1}{2} = {r=3}{},
  cell{1}{3} = {c=8}{},
  cell{2}{3} = {c=7}{c},
  cell{4}{2} = {c},
  cell{4}{3} = {c},
  cell{4}{4} = {c},
  cell{4}{5} = {c},
  cell{4}{6} = {c},
  cell{4}{7} = {c},
  cell{4}{8} = {c},
  cell{4}{9} = {c},
  cell{4}{10} = {c},
  cell{5}{2} = {c},
  cell{5}{3} = {c},
  cell{5}{4} = {c},
  cell{5}{5} = {c},
  cell{5}{6} = {c},
  cell{5}{7} = {c},
  cell{5}{8} = {c},
  cell{5}{9} = {c},
  cell{5}{10} = {c},
  cell{6}{2} = {c},
  cell{6}{3} = {c},
  cell{6}{4} = {c},
  cell{6}{5} = {c},
  cell{6}{6} = {c},
  cell{6}{7} = {c},
  cell{6}{8} = {c},
  cell{6}{9} = {c},
  cell{6}{10} = {c},
  cell{7}{2} = {c},
  cell{7}{3} = {c},
  cell{7}{4} = {c},
  cell{7}{5} = {c},
  cell{7}{6} = {c},
  cell{7}{7} = {c},
  cell{7}{8} = {c},
  cell{7}{9} = {c},
  cell{7}{10} = {c},
  cell{8}{2} = {c},
  cell{8}{3} = {c},
  cell{8}{4} = {c},
  cell{8}{5} = {c},
  cell{8}{6} = {c},
  cell{8}{7} = {c},
  cell{8}{8} = {c},
  cell{8}{9} = {c},
  cell{8}{10} = {c},
  cell{9}{2} = {c},
  cell{9}{3} = {c},
  cell{9}{4} = {c},
  cell{9}{5} = {c},
  cell{9}{6} = {c},
  cell{9}{7} = {c},
  cell{9}{8} = {c},
  cell{9}{9} = {c},
  cell{9}{10} = {c},
  cell{10}{2} = {c},
  cell{10}{3} = {c},
  cell{10}{4} = {c},
  cell{10}{5} = {c},
  cell{10}{6} = {c},
  cell{10}{7} = {c},
  cell{10}{8} = {c},
  cell{10}{9} = {c},
  cell{10}{10} = {c},
  cell{11}{2} = {c},
  cell{11}{3} = {c},
  cell{11}{4} = {c},
  cell{11}{5} = {c},
  cell{11}{6} = {c},
  cell{11}{7} = {c},
  cell{11}{8} = {c},
  cell{11}{9} = {c},
  cell{11}{10} = {c},
  cell{12}{2} = {c},
  cell{12}{3} = {c},
  cell{12}{4} = {c},
  cell{12}{5} = {c},
  cell{12}{6} = {c},
  cell{12}{7} = {c},
  cell{12}{8} = {c},
  cell{12}{9} = {c},
  cell{12}{10} = {c},
  cell{13}{2} = {c},
  cell{13}{3} = {c},
  cell{13}{4} = {c},
  cell{13}{5} = {c},
  cell{13}{6} = {c},
  cell{13}{7} = {c},
  cell{13}{8} = {c},
  cell{13}{9} = {c},
  cell{13}{10} = {c},
  cell{14}{2} = {c},
  cell{14}{3} = {c},
  cell{14}{4} = {c},
  cell{14}{5} = {c},
  cell{14}{6} = {c},
  cell{14}{7} = {c},
  cell{14}{8} = {c},
  cell{14}{9} = {c},
  cell{14}{10} = {c},
  vline{2-3} = {4-14}{},
  hline{1,4-5,10,15} = {-}{},
  hline{3} = {3-9}{},
}
\textbf{Model } & \textbf{F1-Macro } & \textbf{Performance Reduction} &              &              &               &              &              &                       &                \\
                &                    & \textbf{Levels}                &              &              &               &              &              &                       &                \\
                &                    & \textbf{1.1}                   & \textbf{1.2} & \textbf{2.1} & \textbf{~2.2} & \textbf{3.1} & \textbf{3.2} & \textbf{Avg}          & \textbf{AugLy} \\
Naïve           & 0.7436             & 6\%                            & 12\%         & 7\%          & 15\%          & 7\%          & 14\%         & 10\%                  & 7\%            \\
10-static       & 0.7331             & 4\%                            & 9\%          & 6\%          & 10\%          & 7\%          & 13\%         & 8\%                   & 10\%           \\
25-static       & 0.6330             & 4\%                            & \textbf{5\%} & \textbf{3\%} & 7\%           & \textbf{3\%} & \textbf{7\%} & \textbf{\textbf{5\%}} & \textbf{5\%}   \\
50-static       & 0.7293             & 4\%                            & 10\%         & 5\%          & 9\%           & 5\%          & 10\%         & 7\%                   & 15\%           \\
75-static       & 0.7282             & \textbf{2\%}                   & 9\%          & \textbf{3\%} & 10\%          & 4\%          & 9\%          & 6\%                   & 9\%            \\
100-static      & 0.6841             & \textbf{2\%}                   & 8\%          & 4\%          & \textbf{9\%}  & 4\%          & 9\%          & 6\%                   & 7\%            \\
10-dynamic      & 0.7234             & 6\%                            & 11\%         & 6\%          & 12\%          & 7\%          & 13\%         & 9\%                   & 11\%           \\
25-dynamic      & 0.7221             & \textbf{2\%}                   & 9\%          & 3\%          & 15\%          & \textbf{2\%} & 11\%         & 7\%                   & \textbf{3\%}   \\
50-dynamic      & 0.7138             & 5\%                            & 9\%          & 4\%          & 10\%          & 4\%          & 11\%         & 7\%                   & 7\%            \\
75-dynamic      & 0.6429             & 3\%                            & 4\%          & 3\%          & 7\%           & 3\%          & 7\%          & 4\%                   & 9\%            \\
100-dynamic     & 0.6489             & \textbf{2\%}                   & \textbf{3\%} & \textbf{1\%} & \textbf{3\%}  & \textbf{2\%} & \textbf{5\%} & \textbf{\textbf{3\%}} & 5\%            
\end{tblr}
}

%% file: table_mbart_constraint_weakness.tex
\centering
\resizebox{\linewidth}{!}{%
\begin{tblr}{
  row{1} = {c},
  row{3} = {c},
  cell{1}{1} = {r=3}{},
  cell{1}{2} = {r=3}{},
  cell{1}{3} = {c=8}{},
  cell{2}{3} = {c=7}{c},
  cell{4}{2} = {c},
  cell{4}{3} = {c},
  cell{4}{4} = {c},
  cell{4}{5} = {c},
  cell{4}{6} = {c},
  cell{4}{7} = {c},
  cell{4}{8} = {c},
  cell{4}{9} = {c},
  cell{4}{10} = {c},
  cell{5}{2} = {c},
  cell{5}{3} = {c},
  cell{5}{4} = {c},
  cell{5}{5} = {c},
  cell{5}{6} = {c},
  cell{5}{7} = {c},
  cell{5}{8} = {c},
  cell{5}{9} = {c},
  cell{5}{10} = {c},
  cell{6}{2} = {c},
  cell{6}{3} = {c},
  cell{6}{4} = {c},
  cell{6}{5} = {c},
  cell{6}{6} = {c},
  cell{6}{7} = {c},
  cell{6}{8} = {c},
  cell{6}{9} = {c},
  cell{6}{10} = {c},
  cell{7}{2} = {c},
  cell{7}{3} = {c},
  cell{7}{4} = {c},
  cell{7}{5} = {c},
  cell{7}{6} = {c},
  cell{7}{7} = {c},
  cell{7}{8} = {c},
  cell{7}{9} = {c},
  cell{7}{10} = {c},
  cell{8}{2} = {c},
  cell{8}{3} = {c},
  cell{8}{4} = {c},
  cell{8}{5} = {c},
  cell{8}{6} = {c},
  cell{8}{7} = {c},
  cell{8}{8} = {c},
  cell{8}{9} = {c},
  cell{8}{10} = {c},
  cell{9}{2} = {c},
  cell{9}{3} = {c},
  cell{9}{4} = {c},
  cell{9}{5} = {c},
  cell{9}{6} = {c},
  cell{9}{7} = {c},
  cell{9}{8} = {c},
  cell{9}{9} = {c},
  cell{9}{10} = {c},
  cell{10}{2} = {c},
  cell{10}{3} = {c},
  cell{10}{4} = {c},
  cell{10}{5} = {c},
  cell{10}{6} = {c},
  cell{10}{7} = {c},
  cell{10}{8} = {c},
  cell{10}{9} = {c},
  cell{10}{10} = {c},
  cell{11}{2} = {c},
  cell{11}{3} = {c},
  cell{11}{4} = {c},
  cell{11}{5} = {c},
  cell{11}{6} = {c},
  cell{11}{7} = {c},
  cell{11}{8} = {c},
  cell{11}{9} = {c},
  cell{11}{10} = {c},
  cell{12}{2} = {c},
  cell{12}{3} = {c},
  cell{12}{4} = {c},
  cell{12}{5} = {c},
  cell{12}{6} = {c},
  cell{12}{7} = {c},
  cell{12}{8} = {c},
  cell{12}{9} = {c},
  cell{12}{10} = {c},
  cell{13}{2} = {c},
  cell{13}{3} = {c},
  cell{13}{4} = {c},
  cell{13}{5} = {c},
  cell{13}{6} = {c},
  cell{13}{7} = {c},
  cell{13}{8} = {c},
  cell{13}{9} = {c},
  cell{13}{10} = {c},
  cell{14}{2} = {c},
  cell{14}{3} = {c},
  cell{14}{4} = {c},
  cell{14}{5} = {c},
  cell{14}{6} = {c},
  cell{14}{7} = {c},
  cell{14}{8} = {c},
  cell{14}{9} = {c},
  cell{14}{10} = {c},
  vline{2-3} = {4-14}{},
  hline{1,4-5,10,15} = {-}{},
  hline{3} = {3-9}{},
}
\textbf{Model } & \textbf{F1-Macro } & \textbf{Performance Reduction} &              &              &              &              &              &                       &                \\
                &                    & \textbf{Levels}                &              &              &              &              &              &                       &                \\
                &                    & \textbf{1.1}                   & \textbf{1.2} & \textbf{2.1} & \textbf{2.2} & \textbf{3.1} & \textbf{3.2} & \textbf{Avg}          & \textbf{AugLy} \\
Naïve           & 0.9568             & 5\%                            & 20\%         & 7\%          & 24\%         & 10\%         & 26\%         & 15\%                  & 9\%            \\
10-static       & 0.9555             & 3\%                            & 7\%          & 3\%          & 8\%          & 3\%          & 10\%         & 6\%                   & 7\%            \\
25-static       & 0.9415             & 2\%                            & 4\%          & 2\%          & 5\%          & 2\%          & 5\%          & 3\%                   & 5\%            \\
50-static       & 0.9537             & \textbf{1\%}                   & 2\%          & \textbf{1\%} & \textbf{3\%} & \textbf{1\%} & \textbf{3\%} & \textbf{\textbf{2\%}} & 7\%            \\
75-static       & 0.9489             & \textbf{1\%}                   & 3\%          & \textbf{1\%} & \textbf{3\%} & \textbf{1\%} & \textbf{3\%} & \textbf{\textbf{2\%}} & 5\%            \\
100-static      & 0.9140             & \textbf{1\%}                   & \textbf{2\%} & \textbf{1\%} & \textbf{3\%} & \textbf{1\%} & \textbf{3\%} & \textbf{\textbf{2\%}} & \textbf{4\%}   \\
10-dynamic      & 0.9475             & 2\%                            & 5\%          & 2\%          & 6\%          & 3\%          & 7\%          & 4\%                   & 5\%            \\
25-dynamic      & 0.9523             & 2\%                            & 4\%          & 3\%          & 6\%          & 4\%          & 6\%          & 4\%                   & 6\%            \\
50-dynamic      & 0.9406             & \textbf{1\%}                   & 4\%          & 2\%          & 5\%          & 2\%          & 5\%          & 3\%                   & \textbf{4\%}   \\
75-dynamic      & 0.9241             & \textbf{1\%}                   & 3\%          & 2\%          & 4\%          & 2\%          & 4\%          & 3\%                   & \textbf{4\%}   \\
100-dynamic     & 0.9140             & \textbf{1\%}                   & \textbf{2\%} & \textbf{1\%} & \textbf{3\%} & \textbf{1\%} & \textbf{3\%} & \textbf{\textbf{2\%}} & \textbf{4\%}   
\end{tblr}
}